%% file: main.tex
\documentclass{fairmeta}

\microtypesetup{expansion=false}

\usepackage{mathtools}
\usepackage{amssymb}
\usepackage{amsthm}
\usepackage{algorithm}
\usepackage{algpseudocode}

\usepackage{newtxtt}
\usepackage[normalem]{ulem}
\usepackage{silence}

\usepackage{wrapfig}         
\usepackage{needspace}
\makeatletter
\patchcmd{\wrong@fontshape}{\@gobbletwo}{}{}{}
\makeatother
\newtheorem{theorem}{Theorem}

\newtheorem{remark1}[theorem]{Remark}

\DeclareRobustCommand{\metaicon}[2]{%
  \makebox[1.35em][c]{\raisebox{-0.18em}{\includegraphics[height=#1]{#2}}}\hspace{0.35em}%
}
\renewcommand\project[1]{\metadata[\metaicon{1.05em}{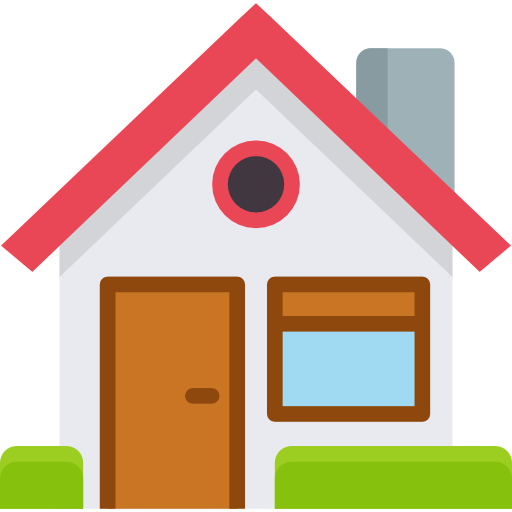}Project]{#1}}
\renewcommand\code[1]{\metadata[\metaicon{1.25em}{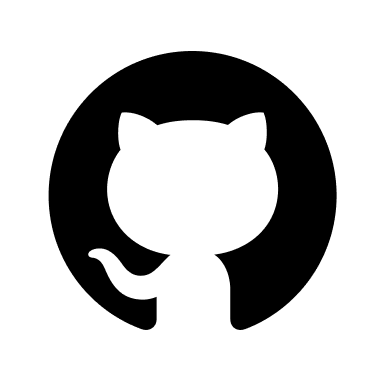}Code]{#1}}
\newcommand\model[1]{\metadata[\metaicon{1.25em}{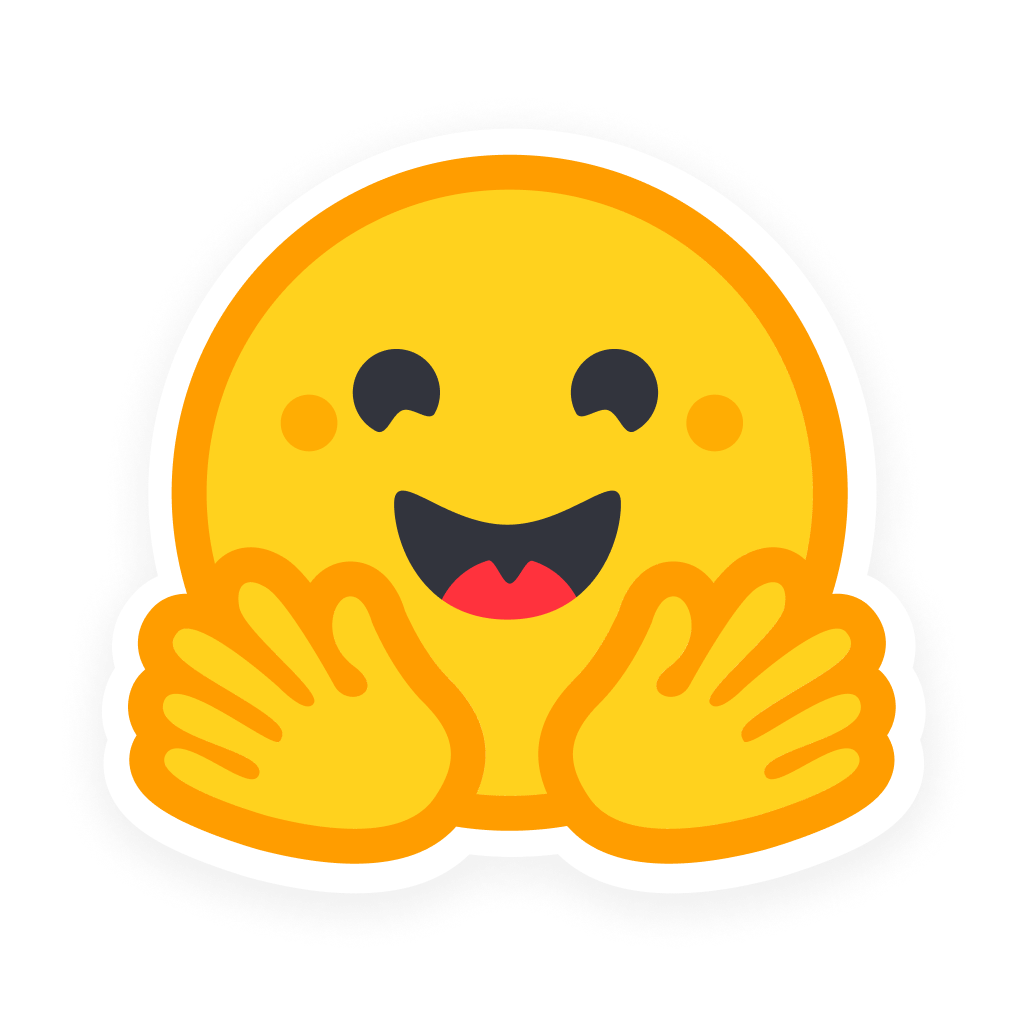}Model]{#1}}

\title{WNM-3D: A World Navigation Model with 3D Scene Conditioning for Closed-Loop VLN}
\author[1,2,3\dagger~~]{\text{Yuehao Huang}}
\author[1,2,4\dagger~~]{\text{Yunzi Wu}}
\author[1,2,5\dagger~~]{\text{Xiaotao Zhang}}
\author[1,2\ddagger]{\text{Xinhai Li}}
\author[1,2]{\text{Jiankun Dong}}
\author[1,6]{\text{Sunyao Zhou}}
\author[3]{\text{Jiajun Lv}}
\author[1,2]{\text{Chi Zhang}}
\author[1,2*]{\text{Chenjia Bai}}
\author[3*]{\text{Yong Liu}}
\author[1,2*]{\text{Xuelong Li}}

\affiliation[1]{Institute of Artificial Intelligence, China Telecom}
\affiliation[2]{Gamma Robotics}
\affiliation[3]{Zhejiang University}
\affiliation[4]{Tongji University}
\affiliation[5]{Shanghai Jiao Tong University}
\affiliation[6]{Fudan University}

\contribution{
\textsuperscript{$\dagger$}Equal Contributions
\quad
\textsuperscript{$\ddagger$}Project Leader
\quad
\textsuperscript{*}Corresponding Authors
}
\date{August, 2026}
\project{\url{https://wnm-3d.github.io}}
\code{\url{https://github.com/TeleHuman/WNM-3D}}
\model{\url{https://huggingface.co/TeleEmbodied/WNM-3D}}
\metadata[Correspondence to] {Yuehao Huang (\email{yuehaohuang@zju.edu.cn}); Chenjia Bai (\email{baicj@chinatelecom.cn})}

\begin{document}
\input{sections/00_abstract}

\maketitle

\input{sections/01_introduction}
\input{sections/02_related_work}
\input{sections/03_method}
\input{sections/04_experiments}

\input{sections/05_conclusion}

\bibliographystyle{unsrtnat}
\bibliography{references} 

\clearpage
\input{sections/06_appendix}

\end{document}

%% file: sections/00_abstract.tex
\abstract{
Recent vision-language navigation (VLN) systems increasingly adapt pretrained vision-language models (VLMs) into vision-language-action (VLA) policies that map egocentric observations and language instructions directly to navigation actions.
Although semantically capable, such action-centric training does not explicitly model how the agent's visual observations evolve under its predicted motion.
Generative world-action models (WAMs) jointly predict future observations and actions, yet existing WAMs for continuous VLN do not condition joint future-view and action generation on geometry-aware representations inferred from the observation history.
We present \textbf{WNM-3D}, a generative \textbf{W}orld \textbf{N}avigation \textbf{M}odel with \textbf{3D} scene conditioning for continuous VLN.
A frozen feed-forward geometry encoder extracts cross-view representations from monocular RGB history, and a trainable \textbf{3D Scene-to-Token Adapter} converts them into a fixed-length scene prefix for the world-action Diffusion Transformer.
Through block-causal attention, this prefix remains visible to every future video-action block throughout joint denoising, providing persistent geometric context.
We train WNM-3D with supervised world-action learning, DAgger-style adaptation on policy-visited states, and Counterfactual DanceGRPO refinement.
Across GN-Bench and VLN-CE, WNM-3D consistently outperforms the backbone-matched WNM-2D counterpart, demonstrating gains from 3D scene conditioning across distinct visual domains.
WNM-3D further outperforms strong VLN baselines on GN-Bench and demonstrates closed-loop deployment on a physical robot.
Component and stage-wise ablations support the Adapter design and the complementary roles of DAgger-SFT and subsequent reward-guided refinement.
}

%% file: sections/01_introduction.tex
\section{Introduction}

Vision-language navigation (VLN) requires an embodied agent to translate language instructions and streaming egocentric observations into navigation actions under partial observability and closed-loop interaction. Recent VLN methods increasingly build action-generating policies upon pretrained vision-language models (VLMs), including video-based VLM policies and explicitly formulated vision-language-action (VLA) models \citep{zhang2024navid,zhang2024uni,cheng2025navila}. These models inherit strong semantic priors for instruction understanding, landmark grounding, and visual reasoning. However, they are optimized primarily through action prediction and therefore do not explicitly model how the agent's observations evolve under its predicted motion. This omission is consequential in continuous navigation, where every action changes the viewpoint and thus the visual evidence available to subsequent decisions. For example, a turn can reveal new regions while moving instruction-relevant landmarks out of view.Action supervision alone provides no direct predictive constraint on this closed-loop observation--action evolution.

Generative world models provide a complementary perspective by predicting how observations evolve through interaction. Action-conditioned navigation world models generate future egocentric observations under candidate actions \citep{bar2025navigation}, whereas generative world-action models (WAMs) jointly model future observations and executable actions within a coupled predictive system \citep{ye2026world,zhao2026worldvln}. Existing WAMs for continuous VLN, however, do not condition joint future-view and action generation on geometry-aware representations inferred from the observation history. RGB histories or video latents preserve appearance and temporal context, but do not directly consolidate multi-view observations into persistent geometry-aware scene context. As a result, the correspondence between predicted motion and its associated visual transition can remain underconstrained, particularly under large viewpoint changes. A monocular observation history nevertheless contains multiple views of the same environment as the agent moves. This motivates our central question: \emph{how can geometry-aware information recovered from this history be incorporated as shared context for both future-view prediction and navigation-action generation?}

We present \textbf{WNM-3D}, a generative \textbf{W}orld \textbf{N}avigation \textbf{M}odel with \textbf{3D} scene conditioning for continuous VLN. To the best of our knowledge, WNM-3D is the first generative world-action model for continuous VLN to incorporate history-derived 3D scene information into joint future-view and action generation. A frozen feed-forward geometry encoder extracts cross-view features from monocular egocentric RGB history, and a trainable \textbf{3D Scene-to-Token Adapter} converts them into a fixed-length prefix aligned with the world-action Diffusion Transformer (DiT). Through block-causal attention, the scene prefix remains visible to every future video-action block throughout joint denoising, providing persistent geometric context. The fixed token interface further decouples the upstream geometry representation from the downstream world-action generator.

Closed-loop navigation exposes the policy to states absent from offline expert demonstrations. We therefore adopt a three-stage curriculum: supervised world-action learning on offline expert trajectories, DAgger-SFT on policy-visited states, and Counterfactual DanceGRPO for reward-guided closed-loop refinement. We evaluate WNM-3D on GN-Bench~\citep{li2026gn0} and VLN-CE~\citep{krantz2020beyond}, which provide high-fidelity 3DGS and reconstruction-based visual domains, respectively, and deploy WNM-3D on a physical robot. WNM-3D consistently outperforms the backbone-matched WNM-2D counterpart on both benchmarks, outperforms VLN baselines on GN-Bench, and demonstrates closed-loop operation in real indoor environments. Component and stage-wise ablations further validate the Adapter design and characterize the complementary roles of DAgger-SFT and DanceGRPO.

Our contributions are threefold:
\begin{itemize}
    \item We formulate a geometry-conditioned WAM for continuous VLN, in which geometry-aware scene tokens derived from the observation history provide shared conditioning for future visual latents and temporally aligned navigation actions.

    \item We introduce a modular 3D Scene-to-Token Adapter that combines geometry-aware feature fusion, content-initialized target queries, anchored deformable resampling, and factorized spatiotemporal refinement to produce a fixed-length prefix compatible with the world-action DiT.

    \item We establish a progressive training curriculum combining supervised world-action learning, DAgger-SFT, and Counterfactual DanceGRPO, with consistent WNM-3D gains on GN-Bench and VLN-CE and closed-loop physical-robot deployment.
\end{itemize}

%% file: sections/02_related_work.tex
\section{Related Work}

\paragraph{VLM-Based Navigation and Closed-Loop Policy Learning.}
Recent navigation methods adapt pretrained vision-language models (VLMs) into action-generating policies, including video-based VLM policies, vision-language-action (VLA) models, and cross-task navigation foundation models \citep{zhang2024navid,zhang2024uni,cheng2025navila,zhang2025embodied,11600818}. Other methods incorporate planning, expert routing, dialogue, scene graphs, spatial memory, or hierarchical reasoning \citep{long2024instructnav,SHENG2026115585,zhou2026deconav,huang2025cogddn,su2026sage,deng2026spacevln,wang2026conavbench,ma2026aura,he2025seeing,wang2023find,wang2024mo,Sheng2026P2DNavPR}. These approaches primarily optimize mappings from observation--instruction histories to actions rather than jointly modeling action-induced future observations. Closed-loop execution further introduces covariate shift; DAgger labels policy-visited states~\citep{ross2011reduction}, while GN0 combines supervised learning, rollout aggregation, and reinforcement learning~\citep{li2026gn0}. Reinforcement learning has also been widely studied for policy optimization and adaptation~\citep{li2024towards,zhang2023fine,zhang2025scrutinize}, while DanceGRPO extends group-relative optimization to diffusion and rectified-flow generators~\citep{xue2025dancegrpo}. At the systems level, device--edge--cloud collaboration has also been explored to alleviate multimodal-model deployment bottlenecks \citep{an2026ai,shao2025ai}.

\paragraph{Generative Navigation World Models and World-Action Models.}
Action-conditioned navigation world models predict future egocentric observations under candidate actions and rely on a separate planner or action selector \citep{bar2025navigation}; SparseVideoNav similarly uses predicted visual futures to guide continuous control \citep{zhang2026sparse}. World-action models (WAMs) instead couple future-state prediction and action generation. DreamZero introduces the joint video-action flow formulation adopted by WNM-3D \citep{ye2026world}. In navigation, WAM-Nav generates latent visual foresight and long-horizon actions \citep{yang2026wam}, while NavWAM models future observations, goal progress, and action chunks in a shared latent sequence \citep{azuma2026navwam}. SWAM jointly generates intermediate RGB-D observations and action trajectories from start and goal RGB inputs, using depth pseudo-labels as spatial supervision \citep{chen2026pondering}; unlike WNM-3D, it does not aggregate an observation history into geometry-aware scene context for joint future-view and action generation. WorldFly and WorldVLN extend generative world-action modeling to aerial navigation \citep{zheng2026worldfly,zhao2026worldvln}. FutureNav jointly optimizes action, dynamics, and future spatial-state objectives but does not formulate future views and actions as coupled generative variables \citep{zhang2026futurenav}. NavWM uses world tokens supervised by depth and semantics for foresight-based trajectory selection~\citep{mei2026navwm}, whereas WNM-3D uses history-derived scene tokens as a shared inference-time condition for jointly generated visual and action variables. Overall, existing navigation WAMs incorporate visual, motion, depth, or spatial cues in different forms, but do not condition joint future-view and action generation on geometry-aware representations inferred from the observation history.

\paragraph{Geometry-Aware World-Action Modeling.}
Feed-forward geometry models such as VGGT and VGGT-$\Omega$ infer camera and dense scene attributes from multiple images \citep{wang2025vggt,wang2026vggt}, while 3DGS~\citep{kerbl3Dgaussians} has been explored for generative 3D modeling and photorealistic robotic simulation \citep{li2023gaussiandiffusion,li2024robogsim}. Geometry has also entered WAMs in other embodied domains: DriveDreamer-Policy jointly models depth, future video, and motion planning for autonomous driving \citep{zhou2026drivedreamer}; GeoSem-WAM uses auxiliary future-geometry and semantic prediction \citep{ma2026geosem}; and MECo-WAM transfers VGGT-derived 4D geometric priors through a training-time geometric expert removed at deployment \citep{zhang2026learning}. These approaches either target non-VLN domains or use geometry mainly through auxiliary objectives or training-time priors. In contrast, WNM-3D converts geometry-aware features inferred from the observation history into scene tokens retained as a shared inference-time condition for future-view and language-conditioned action generation.

%% file: sections/03_method.tex
\section{Method}
\label{sec:method}

\paragraph{Overview.}
WNM-3D jointly predicts finite-horizon future views and temporally aligned navigation actions at each replanning step. As shown in Fig.~\ref{fig:overview}, it builds on the joint video-action flow backbone of DreamZero~\citep{ye2026world} and introduces a frozen geometry encoder followed by a trainable 3D Scene-to-Token Adapter to construct a geometry-aware conditioning prefix from the observation history. \textbf{WNM-2D} shares the same world-action backbone, prediction targets, block-causal attention, and three-stage training curriculum, but replaces this prefix with the backbone's native VAE-encoded RGB history.

\input{figures/overview}

\subsection{Problem Formulation}
\label{sec:problem_formulation}

We consider continuous VLN under partial observability. At replanning step $t$, the agent receives a language instruction $\ell$ and maintains a history $\mathcal{H}_t=(I_{\tau_1},\ldots,I_{\tau_K})$ of $K$ monocular egocentric RGB observations, where $\tau_1<\cdots<\tau_K=t$. The model predicts $B$ future visual blocks $\mathcal{Y}_t=(Y_{t,1},\ldots,Y_{t,B})$, each containing $N_v$ future RGB frames, together with temporally aligned action blocks $\mathcal{A}_t=(\mathcal{A}_{t,1},\ldots,\mathcal{A}_{t,B})$, each containing $N_a$ navigation actions. The current observation $I_t$ immediately precedes the first predicted visual block.

For each replanning step, the simulator trajectory is smoothed and transformed into a fixed local navigation frame anchored at the camera pose corresponding to $I_t$. Let $\mathbf{p}_j^{(t)}=(p_{j,x}^{(t)},p_{j,y}^{(t)})$ denote its $j$-th sampled planar position, with $\mathbf{p}_0^{(t)}=\mathbf{0}$, and let $\psi_j^{(t)}$ denote the corresponding tangent heading. These quantities are used only to construct action targets and are not observed by the policy. Each navigation action is the incremental motion between two consecutive samples:
\begin{equation}
    a_j^{(t)}
    =
    \left(
        p_{j+1,x}^{(t)}-p_{j,x}^{(t)},
        p_{j+1,y}^{(t)}-p_{j,y}^{(t)},
        \operatorname{wrap}_{[-\pi,\pi)}
        \left(\psi_{j+1}^{(t)}-\psi_j^{(t)}\right)
    \right),
    \qquad j=0,\ldots,BN_a-1.
    \label{eq:physical_action}
\end{equation}
Thus, $a_j^{(t)}=(\Delta x_j^{(t)},\Delta y_j^{(t)},\Delta\psi_j^{(t)})$ represents the incremental translation and heading change between consecutive samples in the local frame anchored at $I_t$. The actions are grouped as
\begin{equation}
    \mathcal{A}_{t,b}
    =
    \left(
        a_{(b-1)N_a}^{(t)},\ldots,
        a_{bN_a-1}^{(t)}
    \right)
    \in \mathbb{R}^{N_a\times 3},
    \qquad b\in\{1,\ldots,B\}.
    \label{eq:action_chunk}
\end{equation}
Each action block spans the temporal interval of its paired visual block. For WNM-3D, we model the joint distribution $p_{\theta,\phi}(\mathcal{Y}_t,\mathcal{A}_t\mid\mathcal{H}_t,\ell)$, where $\theta$ parameterizes the world-action backbone and its modality-specific layers, and $\phi$ parameterizes the 3D Scene-to-Token Adapter. WNM-2D follows the same predictive formulation using the native RGB-history prefix and does not include the adapter parameters $\phi$.

\subsection{History Conditioning and Block-Causal Attention}
\label{sec:history_conditioning}

Let $\mathbf{C}_t\in\mathbb{R}^{N_c\times d}$ denote the clean history-conditioning prefix prepended to the self-attention stream of the world-action Diffusion Transformer (DiT), where $N_c$ is the fixed prefix length and $d$ is the DiT hidden width. As illustrated in Fig.~\ref{fig:overview}(a), the two variants differ in how this prefix is constructed: WNM-2D forms $\mathbf{C}_t^{\mathrm{2D}}$ from the backbone's native VAE-encoded RGB history, whereas WNM-3D applies a frozen feed-forward geometry encoder $E_{\mathrm{geo}}$ followed by the trainable 3D Scene-to-Token Adapter $T_\phi$:
\begin{equation}
    \mathbf{C}_t^{\mathrm{3D}}
    =
    T_\phi\!\left(E_{\mathrm{geo}}(\mathcal{H}_t)\right)
    \in \mathbb{R}^{N_c\times d}.
    \label{eq:history_condition}
\end{equation}

In parallel, the current frame and language instruction are encoded through the backbone's native cross-attention pathway as $\mathbf{G}_t=[E_{\mathrm{img}}(I_t);E_{\ell}(\ell)]$. Following DreamZero, the video VAE latent of $I_t$ defines the block-zero visual flow variable $\mathbf{V}_{\sigma_0,0}$.

During training, the current-frame visual group $\mathbf{V}_{\sigma_0,0}$, the future visual groups $\{\mathbf{V}_{\sigma_b,b}\}_{b=1}^{B}$, and the aligned action groups $\{\mathbf{A}_{\sigma_b,b}\}_{b=1}^{B}$ are serialized with the clean prefix as
\begin{equation}
    \mathcal{S}_{\boldsymbol{\sigma}}
    =
    \left[
        \mathbf{C}_t,
        \mathbf{V}_{\sigma_0,0},
        \mathbf{V}_{\sigma_1,1},\ldots,\mathbf{V}_{\sigma_B,B},
        \mathbf{A}_{\sigma_1,1},\ldots,\mathbf{A}_{\sigma_B,B}
    \right].
    \label{eq:denoising_sequence}
\end{equation}
Each future visual-action pair shares the same flow timestep $\sigma_b$ but uses independently sampled noise, while $\mathbf{C}_t$ and $\mathbf{G}_t$ remain clean. The two variants otherwise share the predicted variables, attention layout, backbone, and objectives.

As illustrated in Fig.~\ref{fig:overview}(c), we apply a block-causal self-attention mask indexed by temporal block rather than serialized token position. The prefix attends only within itself, and the block-zero visual group attends to the prefix and itself. For temporal block $b$, its visual and action tokens may attend to the prefix, block zero, and visual-action groups up to and including block $b$, while later blocks remain masked. Visual and action tokens therefore interact bidirectionally within each block, whereas dependencies across blocks remain causal.

\subsection{Joint World-Action Flow Matching}
\label{sec:flow_matching}

As illustrated in Fig.~\ref{fig:overview}(b), we adopt DreamZero's joint flow-matching formulation~\citep{ye2026world} with a shared world-action DiT. Let $\widetilde{\mathcal{Y}}_t=(I_t,Y_{t,1},\ldots,Y_{t,B})$ denote the complete visual sequence, which the frozen video VAE encodes into blockwise latents $\{\mathbf{x}_{\mathrm{data}}^{v,b}\}_{b=0}^{B}$. Action rollouts are scaled, normalized using training-set statistics, and zero-padded to the backbone's fixed action width, yielding $\{\mathbf{x}_{\mathrm{data}}^{a,b}\}_{b=1}^{B}$.

For modality $m\in\{v,a\}$ and temporal block $b$, we sample $\boldsymbol{\epsilon}^{m,b}\sim\mathcal{N}(\mathbf{0},\mathbf{I})$ and construct the linear conditional-flow path
\begin{equation}
    \mathbf{x}_{\sigma_b}^{m,b}
    =
    (1-\sigma_b)\mathbf{x}_{\mathrm{data}}^{m,b}
    +
    \sigma_b\boldsymbol{\epsilon}^{m,b},
    \qquad
    \mathbf{u}_{\sigma_b}^{m,b}
    =
    \boldsymbol{\epsilon}^{m,b}
    -
    \mathbf{x}_{\mathrm{data}}^{m,b}.
    \label{eq:flow_path}
\end{equation}
Visual blocks use $b\in\{0,\ldots,B\}$ and action blocks $b\in\{1,\ldots,B\}$; conditioned on $\mathcal{S}_{\boldsymbol{\sigma}}$, the flow timesteps, and $\mathbf{G}_t$, the shared DiT predicts modality-specific velocity fields. We optimize the world-action objective
\begin{equation}
    \mathcal{L}_{\mathrm{WA}}
    =
    \mathbb{E}
    \left[
        \operatorname{MSE}_{w}
        \left(
            \widehat{\mathbf{u}}_{\theta,\phi}^{v},
            \mathbf{u}^{v}
        \right)
        +
        \lambda_a
        \operatorname{MSE}_{w,\mathbf{M}_a}
        \left(
            \widehat{\mathbf{u}}_{\theta,\phi}^{a},
            \mathbf{u}^{a}
        \right)
    \right],
    \label{eq:world_action_loss}
\end{equation}
where $\widehat{\mathbf{u}}^{m}$ and $\mathbf{u}^{m}$ collect blockwise velocity predictions and targets, and $\lambda_a$ weights the action loss. $\operatorname{MSE}_{w}$ applies blockwise flow-timestep weighting, while $\operatorname{MSE}_{w,\mathbf{M}_a}$ additionally masks padded action dimensions with $\mathbf{M}_a$. The two modalities retain separate input/output transformations and velocity targets while interacting through the shared DiT.

\subsection{3D Scene-to-Token Adapter}
\label{sec:scene_adapter}

The frozen VGGT-$\Omega$~\citep{wang2026vggt} encoder produces $L$ patch-feature grids $\{\mathbf{F}_t^{(l)}\}_{l=1}^{L}$, each of shape $K\times H_s\times W_s\times d_l$. As shown in Fig.~\ref{fig:scene_to_token_adapter}, the adapter maps them to a fixed-length DiT prefix through multi-level fusion, structured query formation, anchored deformable resampling, and factorized spatiotemporal refinement.

\paragraph{Geometry Encoding.}
Each selected encoder level is normalized and projected to a common width $d_s$. Because the feature grids share the same history--height--width lattice, aligned locations can be fused directly without resampling. A lightweight gating network assigns location-adaptive weights across levels, producing the geometry-aware source memory $\bar{\mathbf{F}}_t$:
\begin{equation}
    \mathbf{P}_t^{(l)}
    =
    P_l\!\left(\mathbf{F}_t^{(l)}\right),
    \qquad
    \alpha_{t,i}^{(l)}
    =
    \frac{\exp\!\left(g(\mathbf{P}_{t,i}^{(l)})\right)}
    {\sum_{r=1}^{L}\exp\!\left(g(\mathbf{P}_{t,i}^{(r)})\right)},
    \qquad
    \overline{\mathbf{F}}_{t,i}
    =
    \sum_{l=1}^{L}\alpha_{t,i}^{(l)}\mathbf{P}_{t,i}^{(l)}.
    \label{eq:multilevel_scene_fusion}
\end{equation}
Here, $i$ indexes the shared history--height--width lattice, and $\overline{\mathbf{F}}_t\in\mathbb{R}^{K\times H_s\times W_s\times d_s}$ denotes the fused geometry-aware source memory.

\paragraph{3D Query Resampling.}
We adaptively pool the source memory onto a target lattice and initialize each target query with pooled scene content, a learned slot, and structured embeddings:
\begin{equation}
    \mathbf{B}_t
    =
    \operatorname{Pool}_{T_c,H_c,W_c}
    \left(\overline{\mathbf{F}}_t\right),
    \qquad
    \mathbf{Q}_t^{0}
    =
    \mathbf{Q}_{\mathrm{slot}}
    +
    P_B(\mathbf{B}_t)
    +
    \mathbf{E}_{\mathrm{struct}}.
    \label{eq:scene_query_initialization}
\end{equation}

The target lattice contains $N_c=T_cH_cW_c$ queries. $\mathbf{Q}_{\mathrm{slot}}$ contains learned target-grid slots, while $\mathbf{E}_{\mathrm{struct}}$ aggregates global, temporal, row, column, target-coordinate, and query-type embeddings.

\input{figures/adapter}

The pooled base preserves coarse context but may lose fine-grained evidence. We therefore apply $R$ anchored deformable resampling layers. Each target query uses a regular anchor on the source history--height--width lattice, predicts query-dependent offsets and aggregation weights, and retrieves a small neighborhood augmented with source-coordinate embeddings. This yields refined queries $\mathbf{Q}_t^{R}$ without global cross-attention over the full source memory.

\paragraph{Spatiotemporal Refinement.}
The resampled queries are reshaped to the target lattice and processed by factorized refinement blocks. Each block applies spatial self-attention within each temporal slice, followed by temporal self-attention at each spatial location and an MLP residual. A detail head projects the refined queries, while a coarse head projects the pooled scene base as a residual:
\begin{equation}
    \mathbf{C}_t^{\mathrm{3D}}
    =
    \operatorname{Flatten}
    \left[
        H_{\mathrm{detail}}
        \left(
            \operatorname{STRefine}_{\phi}
            (\mathbf{Q}_t^{R})
        \right)
        +
        H_{\mathrm{coarse}}
        (\mathbf{B}_t)
    \right]
    \in
    \mathbb{R}^{N_c\times d}.
    \label{eq:scene_to_token_adapter}
\end{equation}
The resulting tokens form the clean prefix in Eq.~\eqref{eq:history_condition} and provide shared scene context for future-view and action generation. The adapter is trained jointly with the world-action components, while the geometry encoder remains frozen.

\subsection{Progressive Training and Receding-Horizon Inference}
\label{sec:closed_loop_training}

We optimize WNM-2D and WNM-3D in three stages, progressively shifting from offline expert supervision to policy-induced state correction and reward-guided refinement, as illustrated in Fig.~\ref{fig:training_stages}.

\paragraph{Stage-I: Offline Expert SFT.}
We optimize $\mathcal{L}_{\mathrm{WA}}$ on offline expert trajectories: A*-generated paths on GN-Bench~\citep{hart1968formal} and dataset-provided reference paths on VLN-CE. Each expert action rollout is paired with the aligned future visual continuation. For WNM-3D, the adapter and world-action backbone are trained while VGGT-$\Omega$ remains frozen.

\paragraph{Stage-II: Closed-Loop DAgger-SFT.}
To mitigate closed-loop covariate shift, we roll out the Stage-I policy and query a benchmark-specific expert at policy-visited states~\citep{ross2011reduction}. GN-Bench uses the A* expert, whereas VLN-CE uses the reference-aware navmesh oracle described in Appendix~\ref{app:vlnce_evaluation}. For each visited state, the expert provides a corrective trajectory and the simulator renders the aligned future visual continuation. These samples form the Stage-II DAgger dataset, on which the model is further optimized with $\mathcal{L}_{\mathrm{WA}}$, exposing it to policy-induced states while retaining expert supervision.

\input{figures/training_stages}

\paragraph{Stage-III: Counterfactual DanceGRPO.}
Starting from Stage-II, we adapt DanceGRPO~\citep{xue2025dancegrpo} to the joint visual-action flow policy. We optimize the policy using current-policy world-action samples under a fixed set of navigation contexts without stepping the interactive simulator; generated views and actions are scored using reference observations, expert trajectories, occupancy maps, and navigation metadata. No auxiliary supervised $\mathcal{L}_{\mathrm{WA}}$ term is used.

To localize outcome credit along the diffusion trajectory, we partition eligible DiT source transitions into noise strata and sample one intervention transition from each stratum. For each conditioning instance and selected transition, we generate a group of counterfactual branches that share their initial visual and action latents and all exogenous SDE noise except at the intervention, where branch-specific noise is injected. Each branch completes the reverse-SDE trajectory for reward evaluation, while gradients are replayed only through the intervened transition. This controlled-noise construction associates terminal outcomes with localized diffusion decisions rather than broadcasting the same reward across the entire denoising trajectory.

We maintain separate reward streams for future-view quality, navigation, and stopping. For each stream $q$, the branch advantage is standardized within its conditioning-instance--stratum group:
\begin{equation}
    A^{q}_{i,k,h}
    =
    \frac{
        R^{q}_{i,k,h}-\mu^{q}_{i,k}
    }{
        \sigma^{q}_{i,k}+\epsilon
    },
    \qquad
    \mu^{q}_{i,k}
    =
    \frac{1}{H}
    \sum_{h'=1}^{H}
    R^{q}_{i,k,h'}.
    \label{eq:counterfactual_grpo_adv}
\end{equation}
Here, $H$ is the group size, $\sigma^q_{i,k}$ the within-group reward standard deviation, and $\epsilon$ a numerical stabilizer. Visual rewards score future-view quality and visual-action consistency, while navigation and stopping rewards evaluate action chunks against trajectory and occupancy information.

Let $r^{v}$ denote the current-to-old transition-density ratio for the visual variables at the intervened transition, and let $r^{a}_{c}$ denote the corresponding ratio for action chunk $c$. Using the standard clipped DanceGRPO surrogate $\ell_{\mathrm{clip}}$, we route visual credit through $r^{v}$ and navigation or stopping credit through the action ratio:
\begin{equation}
    \begin{aligned}
        \mathcal{L}_{v}
        &=
        \mathbb{E}
        \left[
            \ell_{\mathrm{clip}}
            \left(
                r^{v},
                A^{v}
            \right)
        \right], \\
        \mathcal{L}_{s,c}
        &=
        \mathbb{E}
        \left[
            \ell_{\mathrm{clip}}
            \left(
                r^{a}_{c},
                A^{s}_{c}
            \right)
        \right],
        \qquad
        s\in\{\mathrm{nav},\mathrm{stop}\}.
    \end{aligned}
    \label{eq:stage3_action_loss}
\end{equation}
Although both modalities update the shared world-action DiT, this modality-routed surrogate prevents reward streams from being indiscriminately assigned across modalities.

The complete Stage-III objective combines the visual, navigation, and stopping losses:
\begin{equation}
    \mathcal{L}_{\mathrm{III}}
    =
    \mathcal{L}_{v}
    +
    \lambda_{\mathrm{act}}
    \left(
        \mathcal{L}_{\mathrm{nav}}
        +
        \lambda_{\mathrm{stop}}
        \mathcal{L}_{\mathrm{stop}}
    \right).
    \label{eq:stage3_total_loss}
\end{equation}
Here, $\mathcal{L}_{\mathrm{nav}}$ and $\mathcal{L}_{\mathrm{stop}}$ denote the chunk-aggregated navigation and stopping losses, while $\lambda_{\mathrm{act}}$ and $\lambda_{\mathrm{stop}}$ weight the action-side and stopping objectives, respectively. Reward definitions and Stage-III optimization details are provided in Appendix~\ref{app:dancegrpo_details}.

\paragraph{Receding-Horizon Inference.}
At replanning step $t$, both variants sample $K$ observations up to $I_t$, recompute the conditioning prefix, and jointly generate finite-horizon visual-action rollouts conditioned on $\mathbf{G}_t$. Only controls derived from the first action block $\mathcal{A}_{t,1}$ are executed before replanning; predicted visual latents are neither decoded into RGB nor fed back into $\mathcal{H}_t$.

%% file: figures/overview.tex
\begin{figure}[t]
    \centering
    \includegraphics[width=\linewidth]{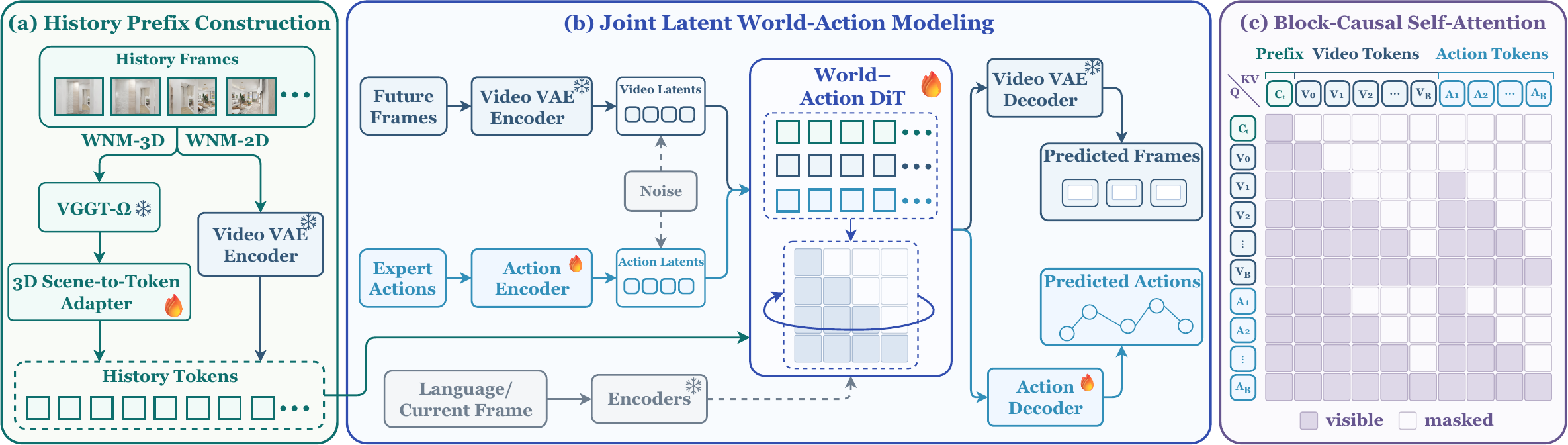}
    \caption{\textbf{Overview of WNM-3D.} 
    (a) \textbf{History prefix construction.} WNM-3D converts history-derived VGGT-$\Omega$ features into a geometry-aware scene prefix, while WNM-2D uses VAE-encoded RGB history.
    (b) \textbf{Joint latent world-action modeling.} The world-action DiT jointly predicts future visual latents and temporally aligned actions conditioned on the history prefix, current frame, and language instruction.
    (c) \textbf{Block-causal self-attention.} Block-causal attention keeps the clean history prefix visible to all future blocks while preserving temporal causality.}
\label{fig:overview}
\end{figure}

%% file: figures/adapter.tex
\begin{wrapfigure}[22]{r}{0.54\linewidth}
    \centering
    \includegraphics[
        width=\linewidth,
        pagebox=cropbox
    ]{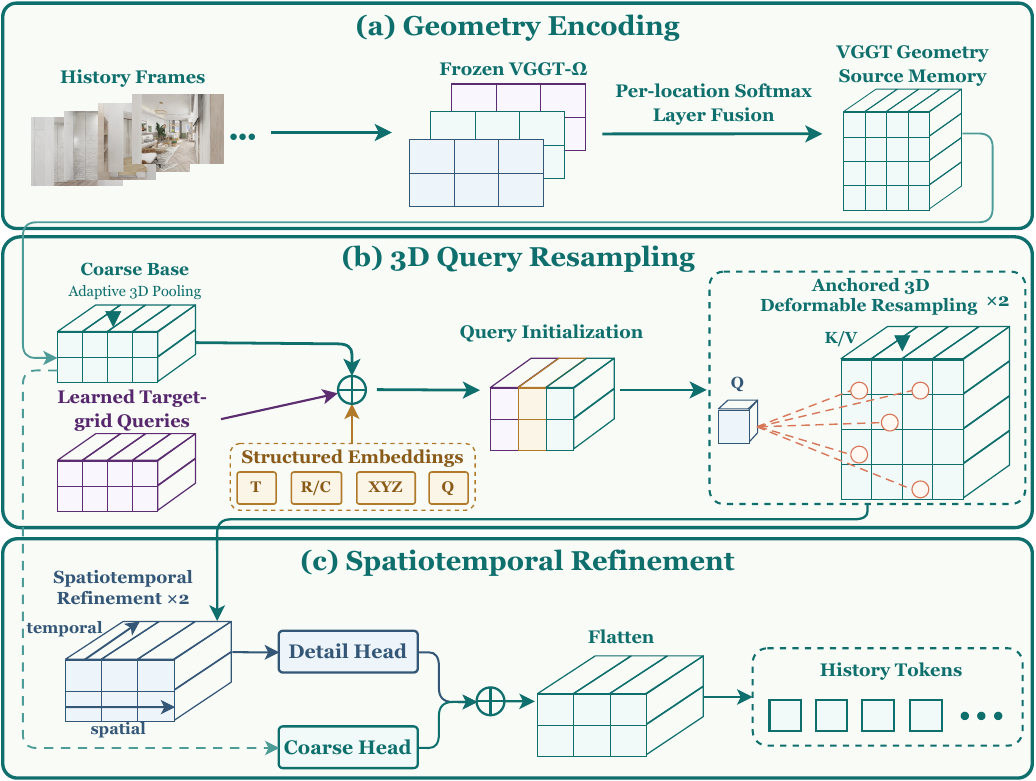}
    \caption{\textbf{3D Scene-to-Token Adapter.} (a) Multi-level VGGT-$\Omega$ features are fused into a geometry-aware source memory. (b) Structured 3D queries resample fine-grained scene features. (c) Spatiotemporal refinement produces fixed-length scene tokens that condition both future-view and action generation.}

    \label{fig:scene_to_token_adapter}
    \vspace{-0.3\baselineskip}
\end{wrapfigure}

%% file: figures/training_stages.tex
\begin{figure}[t]
    \centering
    \includegraphics[width=\linewidth]{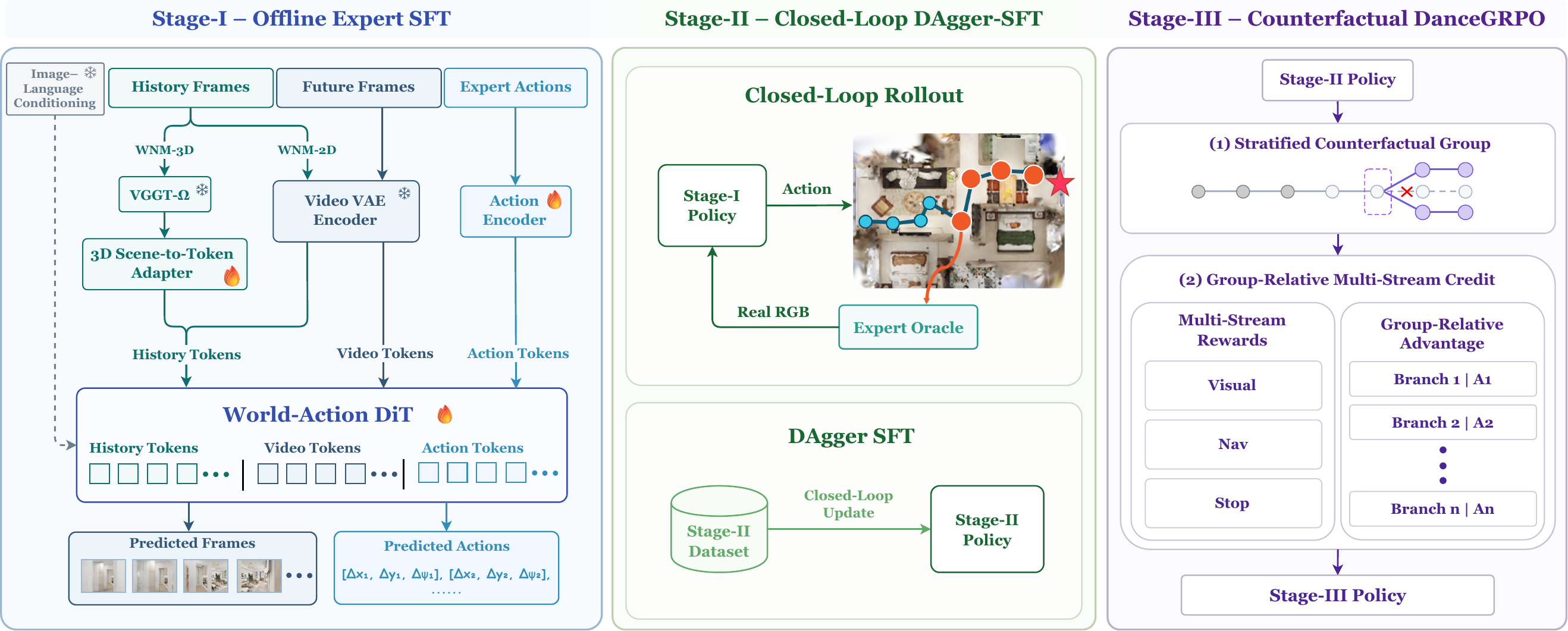}
    \caption{\textbf{Overview of the three-stage training curriculum.}
    (a) \textbf{Stage-I: Offline Expert SFT.} WNM-3D and WNM-2D learn future-view and action generation from offline expert trajectories with geometry-aware and RGB-history conditioning, respectively.
    (b) \textbf{Stage-II: Closed-loop DAgger-SFT.} Expert corrections collected at policy-visited states form a policy-specific DAgger dataset for supervised refinement.
    (c) \textbf{Stage-III: Counterfactual DanceGRPO.} The Stage-II policy is refined using stratified counterfactual world-action groups and group-relative visual, navigation, and stopping rewards.}
    \label{fig:training_stages}
\end{figure}

%% file: sections/04_experiments.tex
\section{Experiments}
\label{sec:experiments}


\subsection{Experimental Setup}
\label{sec:experimental_setup}

\paragraph{Benchmarks and metrics.}
We evaluate WNM-2D and WNM-3D on GN-Bench Seen/Unseen and the VLN-CE R2R Val-Unseen split, which provide high-fidelity 3DGS and reconstruction-based visual domains, respectively. Both WNM policies receive only language and monocular egocentric RGB at inference. We report Navigation Error (NE), Oracle Success (OS), Success Rate (SR), Trajectory Length (TL), and Success weighted by Path Length (SPL) on GN-Bench and NE, OS, SR, and SPL on VLN-CE. Lower NE and higher OS, SR, and SPL are better. Detailed evaluation protocols are provided in Appendix~\ref{app:evaluation_protocol}.

\paragraph{Baselines.} 
On GN-Bench, we compare with CMA~\citep{krantz2020beyond}, NaVid~\citep{zhang2024navid}, UniNaVid~\citep{zhang2024uni}, InternNav(S2)~\citep{internvla-n1}, and GN-BAE~\citep{li2026gn0}, including GN-Matrix-supervised variants reported by GN0 where available. On VLN-CE, we compare with NaVid, NaVILA~\citep{cheng2025navila}, UniNaVid, StreamVLN~\citep{wei2025streamvln}, DualVLN~\citep{wei2026ground}, CorrectNav~\citep{yu2026correctnav}, and GN-BAE. External baselines follow their published training settings and are not strictly data-matched. WNM-2D is our backbone-matched counterpart, replacing the geometry-aware prefix with VAE-encoded RGB history while retaining the same backbone, targets, attention layout, objective, three-stage training protocol, and inference scheme.

\paragraph{Implementation details.}
Both variants initialize from Wan2.2-TI2V-5B, use $K=33$ history frames, and predict $B=4$ blocks with $N_a=8$ actions each. On GN-Bench, Stage-I uses 16K A* demonstrations, Stage-II collects policy-specific DAgger data over the same tasks, and Stage-III starts from the Stage-II checkpoint; VLN-CE uses R2R and RxR training splits with the same curriculum. For WNM-3D, VGGT-$\Omega$ remains frozen while the Adapter and world-action backbone are optimized. Details are in Appendices~\ref{app:implementation_details} and~\ref{app:vlnce_evaluation}.

\subsection{Closed-Loop Navigation}
\label{sec:closed_loop_navigation}

\subsubsection{GN-Bench}
\label{sec:gn_bench_results}

Table~\ref{tab:GN-Bench} reports GN-Bench results. Both WNM variants use only monocular RGB at inference; WNM-3D internally derives geometry-aware scene tokens without depth, BEV observations, or an explicit metric map. Additional rollouts are in Appendix~\ref{app:gnbench_qualitative}.

\input{tables/GN-Bench}

\paragraph{Comparison with prior methods.}
On Seen environments, WNM-3D improves SR and SPL over the strongest prior method in Table~\ref{tab:GN-Bench}, which uses BEV observations, by 30.1 and 22.3 points, respectively. On Unseen environments, it improves over the strongest FPV-only prior baseline by 14.6 SR and 9.3 SPL points.

\paragraph{Effect of 3D scene conditioning.}
WNM-3D improves over WNM-2D by 13.1 SR and 8.0 SPL points on Seen environments and by 7.6 SR and 3.8 SPL points on Unseen environments. This backbone-matched comparison supports the benefit of geometry-aware history conditioning. The gains are larger on Seen, however, and do not indicate a reduced Seen-to-Unseen performance gap.

\FloatBarrier

\subsubsection{VLN-CE}
\label{sec:vlnce_results}

\input{tables/VLN-CE}

Table~\ref{tab:VLN-CE} compares monocular-RGB continuous VLN configurations on R2R Val-Unseen. WNM-3D achieves 4.35 NE, 66.7\% OS, 63.1\% SR, and 57.9\% SPL, yielding competitive performance among methods. Unlike the high-fidelity 3DGS observations in GN-Bench, VLN-CE relies on reconstruction-based observations with artifacts and variations in lighting and scene details~\citep{krantz2022sim}, providing a complementary evaluation under a substantially different visual domain.

\paragraph{Effect of 3D scene conditioning.}
Relative to WNM-2D, WNM-3D improves OS, SR, and SPL by 11.5, 13.0, and 12.6 points, respectively, while reducing NE by 0.99 m. Together with GN-Bench, these gains show that the benefit of geometry-aware history conditioning persists across the two visual domains.

\subsection{Real-World Deployment}
\label{sec:real_world_deployment}

We further deploy WNM-3D on a physical robot to qualitatively assess its closed-loop operation beyond simulation. The robot follows language instructions from live monocular RGB observations across multiple indoor environments using a real-world-adapted checkpoint and the same high-level receding-horizon action-generation and policy-stopping mechanism used in simulation. Figure~\ref{fig:real_world} shows representative rollouts involving multi-step navigation, turning, landmark approach, and stopping. Platform and adaptation details are provided in Appendix~\ref{app:real_world}.

\input{figures/real_world}
 
\subsection{Adapter Component Ablation}
\label{sec:adapter_ablation}

\input{tables/Adapter-Ablation}

To isolate Adapter architecture from subsequent policy adaptation and reward-guided refinement, all variants use only Stage-I Offline Expert SFT with the same frozen VGGT-$\Omega$ encoder, world-action backbone, data, optimization settings, and GN-Bench Unseen evaluation. \emph{Pool + Linear} retains only deepest-level pooling and projection, while the four single-component variants remove multi-level fusion, content initialization, deformable resampling, or spatiotemporal refinement, respectively.

\paragraph{Results.}
Removing any major Adapter component degrades Stage-I Unseen performance. \emph{Pool + Linear} performs worst, indicating that direct pooling and projection underuse the VGGT-$\Omega$ representation. Each single-component ablation also reduces SR and SPL, supporting complementary roles for cross-level fusion, scene-conditioned query initialization, fine-grained resampling, and spatiotemporal refinement.

\subsection{Training-Stage Ablation}
\label{sec:training_stage_ablation}

\input{tables/Training-Stage-Ablation}

We compare Stage-I, Stage-II, the full Stage-III curriculum, and a direct Stage-I$\rightarrow$DanceGRPO variant that skips DAgger-SFT but uses records from the same Stage-I-policy collection as the DAgger dataset. All configurations use the same receding-horizon inference procedure and GN-Bench evaluation episodes.

\paragraph{Effect of DAgger-SFT.}
For WNM-3D, Stage-II improves Seen SR/SPL by 24.0/12.2 points and Unseen SR by 6.6 points, but Unseen SPL drops 3.4 points as TL increases from 5.9 to 13.2. Thus, supervised correction at policy-induced states substantially improves success but does not alone ensure path efficiency under scene-level distribution shift.

\paragraph{Effect of DanceGRPO.}
From Stage-II, DanceGRPO improves WNM-3D by 8.0 SR / 14.3 SPL points on Seen and 3.8 / 8.4 points on Unseen, while reducing TL from 13.9 to 10.6 and 13.2 to 10.2, respectively. WNM-2D shows the same trend (+4.7/+6.7 Seen and +2.2/+3.7 Unseen), indicating improved success and path efficiency after DAgger adaptation.

\paragraph{Role of DAgger-SFT before DanceGRPO.}
Direct Stage-I$\rightarrow$DanceGRPO degrades SR for both variants (WNM-3D: 56.7$\rightarrow$47.1 Seen, 43.1$\rightarrow$39.2 Unseen; WNM-2D: 38.2$\rightarrow$34.1, 35.6$\rightarrow$31.8). Because it and DAgger-SFT start from the same Stage-I checkpoint and draw from the same Stage-I-policy collection, their contrast compares reward-guided refinement with supervised correction under a common source of policy-induced states, although objectives and sample budgets differ. The full pipeline then recollects Stage-III records from the Stage-II policy. We hypothesize that supervised correction broadens feasible policy behavior before group-relative optimization; verifying this mechanism requires direct measurements of within-group action diversity and reward spread.

%% file: tables/GN-Bench.tex
\begin{table}[h]
    \caption{\textbf{Evaluation Metrics on GN-Bench.} In the observation space, \textbf{Depth} denotes depth maps, \textbf{BEV} refers to Bird's Eye View projections, and \textbf{FPV} represents first-person RGB images. \textbf{Bold} represents the best results. \uline{Underline} indicates the second best results. $^{\dagger}$ indicates methods fine-tuned via SFT on the GN-Matrix dataset.}
    \label{tab:GN-Bench}
    \centering
    \setlength{\tabcolsep}{4.35pt}
    \begin{NiceTabular}{lccccccccccccc}
        \CodeBefore
            \columncolor[HTML]{EFEFEF}{2-4}
            \rowcolor[HTML]{F0F8FF}{12-13}
        \Body
        \toprule
        \multirow{2}{*}{\textbf{Method}} & \multicolumn{3}{c}{\textbf{Observation}} & \multicolumn{5}{c}{\textbf{Metrics on Seen}} & \multicolumn{5}{c}{\textbf{Metrics on Unseen}} \\ 
        \cmidrule(lr){2-4} \cmidrule(lr){5-9} \cmidrule(lr){10-14}
         & \textbf{Depth} & \textbf{BEV} & \textbf{FPV} & \textbf{TL} & \textbf{NE} $\downarrow$ & \textbf{OS} $\uparrow$ & \textbf{SR} $\uparrow$ & \textbf{SPL} $\uparrow$ & \textbf{TL} & \textbf{NE} $\downarrow$ & \textbf{OS} $\uparrow$ & \textbf{SR} $\uparrow$ & \textbf{SPL} $\uparrow$ \\ 
        \midrule
        CMA & \checkmark &  & \checkmark & 2.9 & 8.3 & 15.7 & 12.5 & 11.9 & 3.0 & 8.1 & 19.6 & 15.5 & 14.9 \\
        NaVid & &  & \checkmark & 3.4 & 7.9 & 20.1 & 14.6 & 12.8 & 3.5 & 7.7 & 20.3 & 14.5 & 12.8 \\
        UniNaVid &  &  & \checkmark & 4.6 & 7.9 & 22.2 & 15.0 & 12.5 & 5.2 & 7.8 & 20.7 & 12.8 & 10.3 \\
        InternNav(S2) &  &  & \checkmark & 3.6 & 7.4 & 23.1 & 18.8 & 17.5 & 3.7 & 7.2 & 26.7 & 22.1 & 20.3 \\
        NaVid$^\dagger$ &  &  & \checkmark & 2.7 & 7.4 & 19.4 & 18.8 & 18.8 & 2.7 & 7.1 & 23.8 & 23.1 & 23.0 \\
        UniNaVid$^\dagger$ &  &  & \checkmark & 3.7 & 7.2 & 24.1 & 22.5 & 21.9 & 5.8 & 7.5 & 23.1 & 20.8 & 20.2 \\
        InternNav(S2)$^\dagger$ &  &  & \checkmark & 2.9 & 7.1 & 22.5 & 22.4 & 22.4 & 2.9 & 6.9 & 24.9 & 24.0 & 23.7 \\
        GN-BAE &  &  & \checkmark & 5.2 & 4.9 & 48.9 & 46.4 & 44.7 & 5.0 & 5.6 & 43.6 & 38.9 & 37.3 \\ 
        GN-BAE &  & \checkmark & \checkmark & 5.2 & 4.3 & 59.3 & 58.6 & 58.6 & 4.0 & 5.8 & 40.2 & 38.5 & 38.2 \\ 
        \midrule
        \textbf{WNM-2D} &  &  & \checkmark & 7.9 & \uline{2.8} & \uline{80.7} & \uline{75.6} & \uline{72.9} & 7.9 & \uline{4.8} & \uline{52.6} & \uline{45.9} & \uline{42.8} \\ 
        \textbf{WNM-3D} &  &  & \checkmark & 10.6 & \textbf{1.9} & \textbf{94.2} & \textbf{88.7} & \textbf{80.9} & 10.2 & \textbf{4.1} & \textbf{62.2} & \textbf{53.5} & \textbf{46.6}  \\ 
        \bottomrule
    \end{NiceTabular}
\end{table}

%% file: tables/VLN-CE.tex
\begin{wraptable}{r}{0.5\linewidth}
    \vspace{-1.2\baselineskip}
    \caption{\textbf{Evaluation on the VLN-CE R2R Val-Unseen split.}
    WNM-2D and WNM-3D follow the same three-stage curriculum and evaluation protocol.}
    \label{tab:VLN-CE}
    \centering
    \small

    \resizebox{\linewidth}{!}{
        \begin{NiceTabular}{lcccc}
        \CodeBefore
            \rowcolor[HTML]{F0F8FF}{10-11}
        \Body

        \toprule

        \multirow{2}{*}{\textbf{Method}}
        & \multicolumn{4}{c}{\textbf{R2R Val-Unseen}} \\

        \cmidrule(lr){2-5}

        & \textbf{NE} $\downarrow$
        & \textbf{OS} $\uparrow$
        & \textbf{SR} $\uparrow$
        & \textbf{SPL} $\uparrow$ \\

        \midrule

        NaVid      & 5.47 & 49.1 & 37.4 & 35.9 \\
        NaVILA     & 5.22 & 62.5 & 54.0 & 49.0 \\
        UniNaVid   & 5.58 & 53.3 & 47.0 & 42.7 \\
        StreamVLN  & 4.98 & 64.2 & 56.9 & 51.9 \\
        DualVLN    & 4.05 & 70.7 & 64.3 & 58.5 \\
        CorrectNav & 4.24 & 67.5 & 65.1 & 62.3 \\
        GN-BAE     & 3.50 & 75.6 & 67.7 & 63.4 \\

        \midrule

        \textbf{WNM-2D} & 5.34 & 55.2 & 50.1 & 45.3 \\
        \textbf{WNM-3D} & 4.35 & 66.7 & 63.1 & 57.9 \\

        \bottomrule
        \end{NiceTabular}
    }
    \vspace{-3\baselineskip}
\end{wraptable}

%% file: figures/real_world.tex
\begin{figure}[h]
    \centering
    \includegraphics[width=\linewidth]{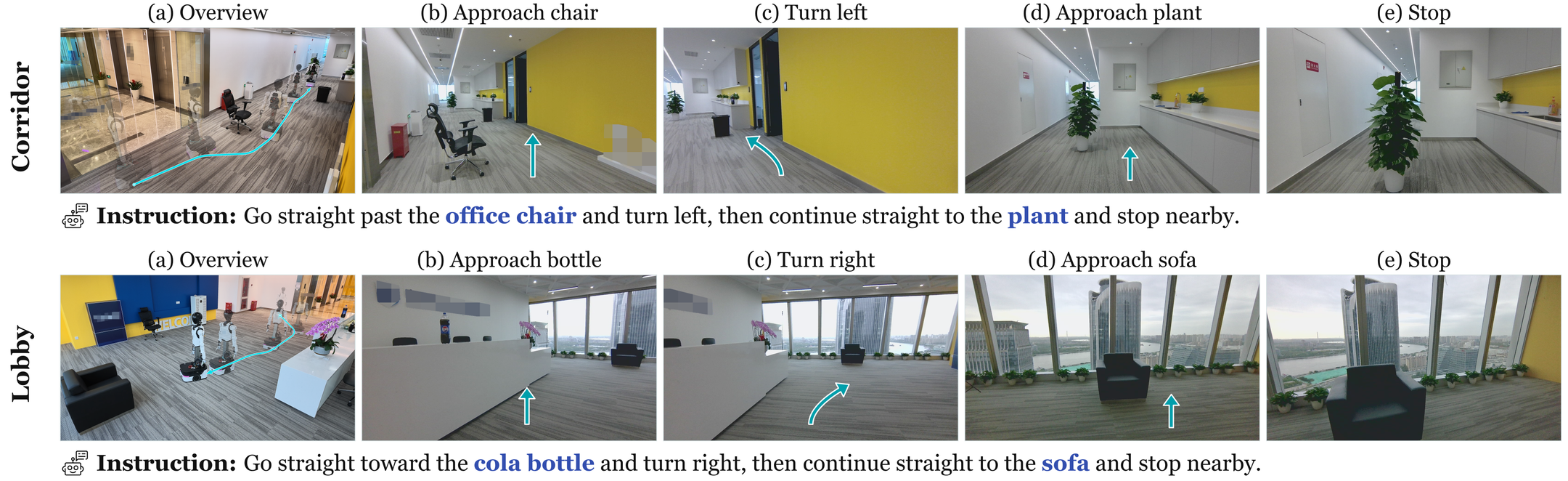}
    \caption{\textbf{Qualitative real-world deployment of WNM-3D.}
    Each row shows a language instruction and selected egocentric observations from a representative closed-loop physical-robot rollout.}
    \label{fig:real_world}
\end{figure}

%% file: tables/Adapter-Ablation.tex
\begin{wraptable}{r}{0.52\linewidth}
    \vspace{-2\baselineskip}
    \caption{\textbf{Ablation of the 3D Scene-to-Token Adapter on GN-Bench Unseen.} All variants are trained only with Stage-I Offline Expert SFT under the same data and optimization settings.}
    \label{tab:adapter_ablation}
    \centering
    \small
    \setlength{\tabcolsep}{9pt}

    \resizebox{\linewidth}{!}{
        \begin{NiceTabular}{lcccc}
        \CodeBefore
            \rowcolor[HTML]{F0F8FF}{8}
        \Body

        \toprule

        \multirow{2}{*}{\textbf{Variant}}
        & \multicolumn{4}{c}{\textbf{GN-Bench Unseen}} \\

        \cmidrule(lr){2-5}

        & \textbf{NE} $\downarrow$
        & \textbf{OS} $\uparrow$
        & \textbf{SR} $\uparrow$
        & \textbf{SPL} $\uparrow$ \\

        \midrule

        Pool + Linear                 & 7.1 & 28.0 & 24.5 & 22.3 \\
        w/o Multi-level Fusion        & 5.7 & 41.5 & 38.2 & 35.1 \\
        w/o Content Initialization    & 6.4 & 37.2 & 35.1 & 32.8 \\
        w/o Deformable Resampling     & 6.9 & 34.9 & 32.0 & 30.1 \\
        w/o Spatiotemporal Refinement & 5.6 & 43.4 & 39.8 & 37.5 \\

        \midrule

        \textbf{Full Adapter} & \textbf{5.0} & \textbf{47.8} & \textbf{43.1} & \textbf{41.6} \\

        \bottomrule
        \end{NiceTabular}
    }
    \vspace{-1.5\baselineskip}
\end{wraptable}

%% file: tables/Training-Stage-Ablation.tex
\begin{table}[h]
    \caption{\textbf{Ablation of the training curriculum for closed-loop navigation on GN-Bench.} Checkmarks $\checkmark$ indicate the stages included in each configuration. Corresponding WNM-2D and WNM-3D configurations use the same downstream backbone and evaluation protocol.}
    \label{tab:training_stage_ablation}
    \centering
    \setlength{\tabcolsep}{2.8pt}
    \begin{NiceTabular}{lccc ccccc ccccc}
        \CodeBefore
            \columncolor[HTML]{EFEFEF}{2-4}
        \Body
        \toprule
        \multirow{2}{*}{\textbf{Model}}
        & \multicolumn{3}{c}{\textbf{Training Stage}}
        & \multicolumn{5}{c}{\textbf{Metrics on Seen}}
        & \multicolumn{5}{c}{\textbf{Metrics on Unseen}} \\
        \cmidrule(lr){2-4} \cmidrule(lr){5-9} \cmidrule(lr){10-14}
        & \textbf{A* SFT}
        & \textbf{DAgger-SFT}
        & \textbf{DanceGRPO}
        & \textbf{TL}
        & \textbf{NE} $\downarrow$
        & \textbf{OS} $\uparrow$
        & \textbf{SR} $\uparrow$
        & \textbf{SPL} $\uparrow$
        & \textbf{TL}
        & \textbf{NE} $\downarrow$
        & \textbf{OS} $\uparrow$
        & \textbf{SR} $\uparrow$
        & \textbf{SPL} $\uparrow$ \\
        \midrule

        \multirow{4}{*}{WNM-3D}
        & \checkmark &            &            & 7.0 & 3.8 & 61.2 & 56.7 & 54.4 & 5.9 & 5.0 & 47.8 & 43.1 & 41.6 \\
        & \checkmark & \checkmark &            & 13.9 & 2.0 & 92.9 & 80.7 & 66.6 & 13.2 & 4.3 & 62.8 & 49.7 & 38.2 \\
        & \checkmark &            & \checkmark & 6.9 & 4.6 & 51.0 & 47.1 & 42.6 & 6.1 & 5.4 & 44.2 & 39.2 & 36.0 \\
        & \checkmark & \checkmark & \checkmark & 10.6 & 1.9 & 94.2 & 88.7 & 80.9 & 10.2 & 4.1 & 62.2 & 53.5 & 46.6 \\

        \midrule

        \multirow{4}{*}{WNM-2D}
        & \checkmark &            &            & 3.9 & 5.2 & 39.9 & 38.2 & 38.0 & 3.7 & 5.6 & 37.2 & 35.6 & 35.3 \\
        & \checkmark & \checkmark &            & 9.0 & 3.0 & 79.4 & 70.9 & 66.2 & 8.9 & 5.0 & 51.9 & 43.7 & 39.1 \\
        & \checkmark &            & \checkmark & 3.6 & 5.6 & 35.7 & 34.1 & 33.8 & 3.4 & 6.0 & 33.7 & 31.8 & 31.4 \\
        & \checkmark & \checkmark & \checkmark & 7.9 & 2.8 & 80.7 & 75.6 & 72.9 & 7.9 & 4.8 & 52.6 & 45.9 & 42.8 \\

        \bottomrule
    \end{NiceTabular}
\end{table}

%% file: sections/05_conclusion.tex
\section{Conclusion}
\label{sec:conclusion}

We presented WNM-3D, a geometry-conditioned generative world-action model for continuous VLN that converts history-derived geometry into a shared scene-token prefix for joint future-view and action generation. WNM-3D consistently outperforms the backbone-matched WNM-2D on GN-Bench and VLN-CE, outperforms strong prior policies on GN-Bench, and demonstrates qualitative closed-loop robot deployment. Component ablations support the 3D Scene-to-Token Adapter design. Training-stage ablations show that DAgger-SFT improves policy adaptation and enables effective subsequent DanceGRPO refinement, which further improves navigation success and path efficiency; applying DanceGRPO directly after Stage-I instead degrades performance.

%% file: sections/06_appendix.tex
\appendix

\section{Implementation Details}
\label{app:implementation_details}

This appendix provides implementation details omitted from the main text, including model configuration, action processing, training-data construction, optimization settings, and the complete Stage-III reward. Benchmark-specific evaluation protocols are provided in Appendix~\ref{app:evaluation_protocol}. Unless stated otherwise, we follow the notation of Sec.~\ref{sec:method}.

\subsection{Architecture and Input Configuration}
\label{app:architecture_details}

WNM-2D and WNM-3D share the same world-action backbone, prediction targets, block-causal attention layout, history-sampling rule, and receding-horizon execution procedure. Their architectural difference is the representation of observed visual context. WNM-2D uses the backbone's native VAE-encoded RGB-history pathway, whereas WNM-3D constructs the clean history-conditioning prefix $\mathbf{C}_t$ with the frozen geometry encoder $E_{\mathrm{geo}}$ and trainable 3D Scene-to-Token Adapter $T_\phi$.

At each replanning step, both variants receive a history $\mathcal{H}_t$ of $K=33$ RGB observations sampled uniformly from the inclusive interval $[I_0,I_t]$. The world-action backbone operates at $160\times320$. WNM-3D additionally resizes the same sampled history to $512\times512$ before applying VGGT-$\Omega$. The prediction horizon comprises $BN_v=32$ future RGB frames and $BN_a=32$ actions, partitioned into $B=4$ aligned blocks with $N_v=8$ frames and $N_a=8$ actions per block.

\input{tables/Appendix-Architecture}

\paragraph{3D Scene-to-Token Adapter configuration.}
For each selected encoder block $l$, the adapter projects $\mathbf{F}_t^{(l)}$ to width $d_s$ and applies the content-adaptive encoder-level fusion in Eq.~\eqref{eq:multilevel_scene_fusion}. The fused source memory is pooled onto a $T_c\times H_c\times W_c$ target lattice to initialize the query content. A structured embedding $\mathbf{E}_{\mathrm{struct}}$---the sum of learned global, temporal, row, and column slots, a Fourier embedding of target coordinates, and a learned query-type embedding---is then added to the linearly projected pooled content. Each of the $R=2$ anchored deformable-resampling layers uses 8 attention heads with 8 sampling points per head. Two factorized spatiotemporal blocks further refine the queries before the detail and coarse branches are projected to the DiT width $d$.

\paragraph{WNM-2D prefix.}
WNM-2D uses the backbone's native VAE-encoded RGB-history prefix and does not include $E_{\mathrm{geo}}$ or $T_\phi$. Its 33 RGB history frames produce 9 VAE latent frames, each contributing a $5\times10$ patch grid. The resulting prefix therefore contains $9\times5\times10=450$ tokens of width $d=3072$, matching the prefix length of WNM-3D. Neither WNM variant receives depth, camera pose, a bird's-eye-view observation, or an explicit metric map at inference time.

\FloatBarrier

\subsection{Action Encoding, Decoding, and Execution}
\label{app:action_execution}

The policy represents the physical increments in Eq.~\eqref{eq:physical_action} using the backbone's fixed-width action vector. Before flow matching, the three physical dimensions are multiplied by $s_a=4$, normalized with training-set action quantiles, and zero-padded to width $d_a=32$. At inference, the inverse transform is applied to the first three predicted dimensions. Let $\widehat{\mathbf{a}}_j^{(t)}\in[-1,1]^3$ denote the unpadded normalized prediction associated with $a_j^{(t)}$. We decode it as
\begin{equation}
    a_j^{(t)}
    =
    \frac{1}{s_a}
    \left[
        \frac{\widehat{\mathbf{a}}_j^{(t)}+1}{2}
        \odot
        \left(
            \mathbf{q}_{99}-\mathbf{q}_{01}
        \right)
        +
        \mathbf{q}_{01}
    \right],
    \qquad
    s_a=4,
    \label{eq:app_action_denormalization}
\end{equation}
where $\mathbf{q}_{01}$ and $\mathbf{q}_{99}$ are the checkpoint-associated $1\%$ and $99\%$ action quantiles.

The action coordinate system is anchored at the camera pose associated with the current observation $I_t$. Simulator pose is used only to transform predicted local increments for execution and reward computation and is never exposed to the policy. For $j\in\{0,\ldots,31\}$, the world position after applying action $j$ is
\begin{equation}
    \mathbf{p}_{j+1}^{\mathrm{world}}
    =
    \mathbf{p}_{0}^{\mathrm{world}}
    +
    R_t
    \sum_{u=0}^{j}
    \left(
        \Delta x_u^{(t)},
        \Delta y_u^{(t)}
    \right),
    \label{eq:app_world_trajectory}
\end{equation}
where $R_t$ is the planar camera-to-navigation rotation at $I_t$. All planar increments are expressed in this fixed local frame; predicted yaw increments are accumulated separately and do not recursively rotate the XY increments.

For chunk $c\in\{1,\ldots,B\}$, let its physical XY path length be
\begin{equation}
L_c^{xy}
=
\sum_{j=(c-1)N_a}^{cN_a-1}
\sqrt{
\left(\Delta x_j^{(t)}\right)^2
+
\left(\Delta y_j^{(t)}\right)^2
}.
\end{equation}
Stage-III stopping credit uses the following ordered STOP detector. If
$L_c^{xy}<0.15\,\mathrm{m}$, STOP is emitted before any action in the chunk is executed. Otherwise, actions are inspected in temporal order, and the first action satisfying
\[
\max
\left(
\left|\Delta x_j^{(t)}\right|,
\left|\Delta y_j^{(t)}\right|,
\left|\Delta\psi_j^{(t)}\right|
\right)
\le 10^{-3}
\]
emits STOP immediately before that action is applied. If neither condition is satisfied, the chunk does not terminate. The path-length comparison is strict, i.e., $<0.15\,\mathrm{m}$ rather than $\le 0.15\,\mathrm{m}$.

During Stage-III reward computation, this detector is evaluated independently for all four action chunks to construct counterfactual chunk-wise stopping credit. At receding-horizon inference, execution is based only on the first predicted action chunk before replanning. Collision-aware reward computation distinguishes the raw generated plan from a deployment-simulated execution of the first chunk. The first chunk is executed against a four-pixel-dilated occupancy map: when a segment contacts the dilated obstacle set, execution is clipped to the last free point and the remaining predicted increments continue from that actual position rather than terminating the chunk. A separate two-pixel-dilated raw-plan check is retained for hard-collision diagnostics and supplies the hard-collision event for later counterfactual chunks. Independently, every raw chunk is scored by a soft-clearance risk over the two-to-six-pixel warning band. Recovery credit is deployment-specific: only the first chunk can receive a bounded recovery bonus, evaluated from its contact-clipped execution using a six-pixel tail-clearance requirement together with escape-distance and post-collision STOP checks. All collision, clearance, and recovery terms remain inside the navigation reward; no separate collision GRPO stream is enabled. Full definitions are provided in Sec.~\ref{app:navigation_reward}, and stopping definitions are given in Sec.~\ref{app:stopping_reward}.

\FloatBarrier

\subsection{Stage-I and Stage-II Training}
\label{app:sft_dagger_details}

Unless otherwise stated, the Stage-I and Stage-II data statistics below correspond to GN-Bench training; the VLN-CE data configuration is described in Appendix~\ref{app:vlnce_evaluation}.

\paragraph{Stage-I: Offline Expert SFT.}
We construct the offline training set from 16K instruction-conditioned trajectories generated by an A* expert. For each sampled trajectory, a valid replanning timestep is selected uniformly, and the observation context and subsequent 32-step visual-action targets are constructed at that timestep. Each WNM variant is trained for 20 epochs with $\mathcal{L}_{\mathrm{WA}}$ from Eq.~\eqref{eq:world_action_loss}.

\paragraph{Stage-II: Closed-loop DAgger-SFT.}
Starting from its Stage-I checkpoint, each WNM variant is rolled out on the same 16K navigation tasks. At every policy-visited state, the A* expert provides a corrective action sequence, and the simulator renders the trajectory-consistent future observations. These policy-induced samples form the Stage-II DAgger datasets, which contain approximately 691K chunks for WNM-2D and 633K chunks for WNM-3D. Each variant is then optimized for 5 epochs with $\mathcal{L}_{\mathrm{WA}}$.

\input{tables/Appendix-SFT-DAgger}

\paragraph{Flow-matching timesteps and loss weighting.}
Stages I and II use the same discrete training schedule with 1,000 indices. For each future block $b$, we sample
\begin{equation}
    \begin{aligned}
        k_b &\sim \operatorname{Uniform}\{0,\ldots,999\},
        &
        \bar{\sigma}_{k_b} &= 1-\frac{k_b}{1000}, \\
        \sigma_b
        &=
        \frac{5\bar{\sigma}_{k_b}}
        {1+4\bar{\sigma}_{k_b}}.
    \end{aligned}
    \label{eq:app_flow_timestep_sampling}
\end{equation}
The two VAE latent frames representing the eight RGB frames in block $b$ and the corresponding eight action steps share $\sigma_b$; their visual and action Gaussian noise samples remain independent. The block-zero latent corresponding to the current target-prefix frame is assigned an independently sampled index from the same schedule.

The timestep-dependent weight in Eq.~\eqref{eq:world_action_loss} is defined on this shifted schedule as
\begin{equation}
    \begin{aligned}
        q_k
        &=
        \exp\!\left[-2\left(\sigma_k-\frac{1}{2}\right)^2\right],
        &
        q_{\min}
        &=
        \min_{0\le r\le999}q_r, \\
        w_k
        &=
        \frac{1000\left(q_k-q_{\min}\right)}
        {\sum_{j=0}^{999}\left(q_j-q_{\min}\right)}.
    \end{aligned}
    \label{eq:app_flow_timestep_weight}
\end{equation}
Thus, the weights have unit mean over the 1,000 training indices. $\operatorname{MSE}_{w}$ denotes the mean elementwise squared velocity error after weighting each block by its corresponding $w_k$. For $\operatorname{MSE}_{w,\mathbf{M}_a}$, the mask $\mathbf{M}_a$ sets the squared residuals of zero-padded action dimensions to zero before the same weighting and reduction are applied. We use equal coefficients for the visual and action terms, i.e., $\lambda_a=1$.

Stages I and II use AdamW with an initial learning rate of $1\times10^{-5}$, $\beta_1=0.95$, $\beta_2=0.999$, $\epsilon=10^{-8}$, weight decay $1\times10^{-5}$, and a cosine schedule with a $5\%$ warm-up and a zero terminal learning rate. Gradients are clipped to a maximum norm of 1.0. Training uses BF16 arithmetic, TF32 matrix multiplication, and 16 NVIDIA H100 GPUs. The global batch size is 256 for WNM-2D and 192 for WNM-3D; the latter is reduced to accommodate geometry-aware history encoding.

\FloatBarrier

\subsection{Stage-III Counterfactual DanceGRPO}
\label{app:dancegrpo_details}

For GN-Bench, Stage-III optimization uses about 96K fixed policy-generated records collected in simulation for each configuration. For the direct A* SFT--DanceGRPO ablation, we use records from the same Stage-I-policy collection used to construct the DAgger-SFT dataset and initialize DanceGRPO directly from the Stage-I checkpoint, without performing the DAgger-SFT update. For the full A* SFT--DAgger-SFT--DanceGRPO curriculum, we instead recollect the Stage-III records using the resulting Stage-II policy and initialize DanceGRPO from the corresponding Stage-II checkpoint. Each record contains simulator-rendered future observations, an expert trajectory, occupancy information, and navigation metadata. These fixed annotations define the visual, navigation, and stopping reward streams; the simulator is not stepped during Stage-III optimization, and no auxiliary supervised $\mathcal{L}_{\mathrm{WA}}$ term is used.

\subsubsection{Denoising-Stratum Counterfactual Sampling}
\label{app:dancegrpo_strata}

Following Sec.~\ref{sec:closed_loop_training}, we partition the eligible denoising transitions into $K_s$ strata $\{\mathcal{S}_k\}_{k=1}^{K_s}$. The final configuration uses $K_s=4$ and
\begin{equation}
    \mathcal{S}_1=\{0,1,2\},
    \qquad
    \mathcal{S}_2=\{6\},
    \qquad
    \mathcal{S}_3=\{10\},
    \qquad
    \mathcal{S}_4=\{13,14,15\},
    \label{eq:app_diffusion_strata}
\end{equation}
where each element is a zero-based transition index in the 16-step sampler. At every optimization step, one transition $\tau_k\in\mathcal{S}_k$ is sampled from each stratum, and the resulting set $\{\tau_k\}_{k=1}^{K_s}$ is shared by the entire condition batch.

For every conditioning instance $i$ and stratum $k$, we generate a group of $H=4$ counterfactual branches $h\in\{1,2,3,4\}$. The four branches share their initial visual and action latents and all exogenous SDE increments except at $\tau_k$, where branch-specific visual and action noise is injected. Each conditioning instance therefore produces $HK_s=4K_s=16$ complete world-action samples. Rewards are evaluated on the completed world-action samples, but gradients are replayed only through the intervened transition $\tau_k$. Visual, navigation, and stopping rewards are standardized independently within each four-branch group according to Eq.~\eqref{eq:counterfactual_grpo_adv}. If all valid branches in a group have identical rewards, their advantages are zero.

The visual and action transitions use independent Gaussian SDE noise factors with stochasticity coefficients
\begin{equation}
    \eta_v=0.70,
    \qquad
    \eta_a=0.20.
    \label{eq:app_sde_coefficients}
\end{equation}
Transition log densities are reduced over their event dimensions before forming the modality-specific likelihood ratios $r^v$ and $r_c^a$. As described in Sec.~\ref{sec:closed_loop_training}, the visual reward is routed through $r^v$, whereas the navigation and stopping rewards of chunk $c$ are routed through $r_c^a$.

For all reward streams, we use the standard clipped surrogate
\begin{equation}
    \ell_{\mathrm{clip}}(r,A;\varepsilon)
    =
    -\min
    \left(
        rA,\,
        \operatorname{clip}
        \left(
            r,1-\varepsilon,1+\varepsilon
        \right)A
    \right),
    \label{eq:app_ppo_clip}
\end{equation}
with modality-specific clipping thresholds given in Table~\ref{tab:app_dancegrpo_hyperparameters}. The action density is reduced only over the three deployed dimensions $(\Delta x,\Delta y,\Delta\psi)$ of each chunk; padded action dimensions do not contribute to the policy ratio.

\subsubsection{Visual Reward}
\label{app:visual_reward}

The visual reward combines a base future-view quality score with a visual-side flow-action consistency bonus:
\begin{equation}
    R^v
    =
    R_{\mathrm{vision\text{-}base}}
    +
    R_{\mathrm{flow}}^v.
    \label{eq:app_visual_reward}
\end{equation}
The base score evaluates the generated future frames in terms of structural similarity, pixel reconstruction, and temporal consistency:
\begin{equation}
    Q_v
    =
    0.60S_{\mathrm{Pyr\text{-}SSIM}}
    +
    0.25
    \left(
        1-E_{\mathrm{charb}}
    \right)
    +
    0.15
    \left(
        1-E_{\mathrm{temp}}
    \right),
    \label{eq:app_visual_quality}
\end{equation}
where $S_{\mathrm{Pyr\text{-}SSIM}}$ is our pyramid-averaged SSIM score. For the $T$ evaluated frames, it is defined as
\begin{equation}
    S_{\mathrm{Pyr\text{-}SSIM}}
    =
    \frac{1}{4T}
    \sum_{m=1}^{4}
    \sum_{t=1}^{T}
    \operatorname{SSIM}
    \left(
        \mathcal{Y}_{t}^{(m)},
        \mathcal{Y}_{t,\mathrm{GT}}^{(m)}
    \right),
    \label{eq:app_pyramid_ssim}
\end{equation}
where scale $m=1$ corresponds to $64\times128$, and each subsequent scale is obtained by $2\times2$ average pooling with stride 2. This score is an arithmetic average of per-scale SSIM values and should not be interpreted as the standard multiplicative MS-SSIM index. $E_{\mathrm{charb}}$ is the Charbonnier reconstruction error, and $E_{\mathrm{temp}}$ is the Charbonnier error between predicted and ground-truth temporal differences. Both error terms use $\epsilon=10^{-3}$ and are clipped to $[0,1]$. A degradation factor $D_{\mathrm{deg}}\in[0,1]$ downweights invalid, frozen, low-contrast, exposure-corrupted, flickering, or excessively dynamic generations:
\begin{equation}
    R_{\mathrm{vision\text{-}base}}
    =
    D_{\mathrm{deg}}Q_v.
    \label{eq:app_visual_base_reward}
\end{equation}
The target-prefix frame is excluded from the future-frame quality and degradation statistics. After checking both sequences for finite values and rejecting predictions with more than $1\%$ of values outside $[0,1]$, the remaining predicted and ground-truth RGB frames are clipped to $[0,1]$ and bilinearly resized with antialiasing to $64\times128$. Let $\widetilde{\mathcal{Y}}^{\mathrm{pred}}_t,\widetilde{\mathcal{Y}}^{\mathrm{GT}}_t\in[0,1]^{3\times64\times128}$ denote these resized future frames. We define the predicted and ground-truth motion magnitudes as the mean absolute RGB change between adjacent future frames,
\begin{equation}
    \begin{aligned}
    M_{\mathrm{pred}}
    &=
    \operatorname*{mean}_{t=2,\ldots,T;\,c,h,w}
    \left|
        \widetilde{\mathcal{Y}}^{\mathrm{pred}}_{t,c,h,w}
        -
        \widetilde{\mathcal{Y}}^{\mathrm{pred}}_{t-1,c,h,w}
    \right|, \\
    M_{\mathrm{GT}}
    &=
    \operatorname*{mean}_{t=2,\ldots,T;\,c,h,w}
    \left|
        \widetilde{\mathcal{Y}}^{\mathrm{GT}}_{t,c,h,w}
        -
        \widetilde{\mathcal{Y}}^{\mathrm{GT}}_{t-1,c,h,w}
    \right|.
    \end{aligned}
    \label{eq:app_visual_motion_statistics}
\end{equation}
For the flicker check, let
\begin{equation}
    \mu_t^{\mathrm{pred}}
    =
    \operatorname*{mean}_{c,h,w}
    \widetilde{\mathcal{Y}}^{\mathrm{pred}}_{t,c,h,w},
    \qquad
    \mu_t^{\mathrm{GT}}
    =
    \operatorname*{mean}_{c,h,w}
    \widetilde{\mathcal{Y}}^{\mathrm{GT}}_{t,c,h,w},
\end{equation}
and define the maximum adjacent-frame mean-intensity changes
\begin{equation}
    \Delta_{\mathrm{int}}^{\mathrm{pred}}
    =
    \max_{t=2,\ldots,T}
    \left|
        \mu_t^{\mathrm{pred}}-\mu_{t-1}^{\mathrm{pred}}
    \right|,
    \qquad
    \Delta_{\mathrm{int}}^{\mathrm{GT}}
    =
    \max_{t=2,\ldots,T}
    \left|
        \mu_t^{\mathrm{GT}}-\mu_{t-1}^{\mathrm{GT}}
    \right|.
    \label{eq:app_visual_intensity_jump}
\end{equation}
Thus, $M_{\mathrm{pred}}$, $M_{\mathrm{GT}}$, $\Delta_{\mathrm{int}}^{\mathrm{pred}}$, and $\Delta_{\mathrm{int}}^{\mathrm{GT}}$ are dimensionless statistics in normalized RGB units; they are not optical-flow magnitudes. Within the future-frame quality and degradation statistics, the excluded target-prefix frame is used only for the separate prefix-copy diagnostic; the flow-action term below additionally uses it to form the first adjacent-frame optical-flow pair. The per-frame RGB spatial standard deviation used by the constant-video and low-contrast checks is computed on the clipped future prediction before the $64\times128$ resize, whereas the mean-intensity and saturation-ratio statistics are computed on the resized future prediction.

We set $D_{\mathrm{deg}}=0$ for non-finite predictions, predictions with more than $1\%$ of values outside $[0,1]$, or predictions whose mean per-frame RGB-value standard deviation is below 0.01. Otherwise, $D_{\mathrm{deg}}$ is initialized to 1.0. It is capped at 0.10 for a frozen video, defined by $M_{\mathrm{GT}}\ge0.01$ and $M_{\mathrm{pred}}<\max(0.002,0.10M_{\mathrm{GT}})$, or for prefix copying, defined by a mean ground-truth prefix change of at least 0.02 and a mean prediction-to-prefix error of at most 0.01. The factor is capped at 0.25 if the prediction has an RGB-value standard deviation below 0.03, a mean RGB intensity outside $[0.03,0.97]$, more than $85\%$ of its values at or below 0.02 or at or above 0.98, $\Delta_{\mathrm{int}}^{\mathrm{pred}}>\max(0.08,3\Delta_{\mathrm{int}}^{\mathrm{GT}})$, or $M_{\mathrm{pred}}>\max(0.12,3M_{\mathrm{GT}})$. If several conditions apply, we use the smallest factor.

Because receding-horizon execution is based on $\mathcal{A}_{t,1}$ before replanning, the flow-action term is evaluated on the generated visual block aligned with $\mathcal{A}_{t,1}$. Let
\begin{equation}
    \Delta\mathbf{p}_{a}
    =
    \sum_{j=0}^{N_a-1}
    \left(
        \Delta x_j^{(t)},
        \Delta y_j^{(t)}
    \right)
    \in\mathbb{R}^{2}
    \label{eq:app_action_displacement}
\end{equation}
denote the cumulative XY displacement of the executed action block. The decoded sequence used for this comparison contains the current target-prefix frame followed by the aligned visual block's $N_v=8$ generated future frames. These nine frames are converted to $64\times128$ grayscale images, and forward and backward DIS flow is computed for the resulting eight adjacent-frame pairs. The target-prefix frame is used only to form the first pair and is excluded from the future-frame visual-quality score. For each pair, we form a descriptor from the median $(u,v)$ flow in each cell of a $4\times6$ grid and the global p50, p75, and p90 flow-magnitude statistics; $\varphi_{\mathrm{flow}}$ is the mean of these eight descriptors. For each Stage-III configuration, a ridge regressor is fitted once on 480 randomly sampled ground-truth clips from its corresponding training records and kept fixed during optimization to map optical-flow descriptors to local camera motion:
\begin{equation}
    \widehat{\mathbf{a}}_{\mathrm{flow}}
    =
    W_{\mathrm{flow}}\varphi_{\mathrm{flow}}
    +
    \mathbf{b}_{\mathrm{flow}}
    \in\mathbb{R}^{3},
    \label{eq:app_flow_calibration}
\end{equation}
which estimates $(\Delta x,\Delta y,\Delta\psi)$. We define $\widehat{\Delta\mathbf{p}}_{\mathrm{flow}}\in\mathbb{R}^{2}$ as its first two components; the yaw component is used only for diagnostics.

For a moving action block, the directional agreement is
\begin{equation}
    s_{xy}
    =
    \max
    \left(
        0,\,
        \frac{
            \Delta\mathbf{p}_{a}^{\mathsf T}
            \widehat{\Delta\mathbf{p}}_{\mathrm{flow}}
        }{
            \|\Delta\mathbf{p}_{a}\|_2
            \|\widehat{\Delta\mathbf{p}}_{\mathrm{flow}}\|_2
        }
    \right),
    \label{eq:app_flow_direction_score}
\end{equation}
provided $\|\Delta\mathbf{p}_{a}\|_2\ge\delta_a$ and $\|\widehat{\Delta\mathbf{p}}_{\mathrm{flow}}\|_2\ge\delta_f$. If the action block is near stationary, $\|\Delta\mathbf{p}_{a}\|_2<\delta_a$, we instead measure whether the generated video is also approximately stationary:
\begin{equation}
    s_{xy}
    =
    \exp
    \left[
        -
        \left(
            \frac{m_{90}}{\delta_{\mathrm{stat}}}
        \right)^2
    \right],
    \label{eq:app_stationary_flow_score}
\end{equation}
where $m_{90}$ is the p90 component of the temporally averaged flow descriptor. When the action requests clear motion but the inferred visual translation is below $\delta_f$, we set $s_{xy}=0$.

For adjacent-frame pair $q$, flow confidence combines forward--backward consistency and image-texture reliability. The video-level confidence is
\begin{equation}
    C_{\mathrm{flow}}
    =
    \operatorname{median}_{q=1,\ldots,8}
    \left(C_{\mathrm{fb},q}C_{\mathrm{tex},q}\right),
    \label{eq:app_flow_confidence}
\end{equation}
where $C_{\mathrm{fb},q}$ is the fraction of consistent pixels in pair $q$. Specifically, for a forward vector $\mathbf{f}(\mathbf{x})$ and backward flow $\mathbf{b}$ sampled at $\mathbf{x}+\mathbf{f}(\mathbf{x})$, a pixel is consistent only if the destination is in bounds and $\|\mathbf{f}(\mathbf{x})+\mathbf{b}(\mathbf{x}+\mathbf{f}(\mathbf{x}))\|_2\le 1.0+0.05\|\mathbf{f}(\mathbf{x})\|_2$ pixels. The texture statistic $g_q$ is the mean magnitude of the $3\times3$ Sobel gradient in the preceding grayscale frame, and $C_{\mathrm{tex},q}=\operatorname{clip}(g_q/8,0,1)$. The final consistency score and reward contribution are
\begin{equation}
    S_{\mathrm{flow-act}}
    =
    s_{xy}C_{\mathrm{flow}},
    \qquad
    R_{\mathrm{flow}}^v
    =
    \lambda_{\mathrm{flow}}^v
    D_{\mathrm{deg}}
    S_{\mathrm{flow-act}},
    \qquad
    \lambda_{\mathrm{flow}}^v=0.05.
    \label{eq:app_flow_action_reward}
\end{equation}
The inferred yaw component does not contribute to $S_{\mathrm{flow-act}}$ or either flow-action reward branch. The same gated score $D_{\mathrm{deg}}S_{\mathrm{flow-act}}$ also provides the action-side first-chunk bonus defined in Appendix~\ref{app:navigation_reward}.

\input{tables/Appendix-Visual-Reward}

\FloatBarrier

\subsubsection{Navigation Reward}
\label{app:navigation_reward}

For each valid action chunk $c\in\{1,\ldots,B\}$, the navigation reward combines signed geodesic progress, path-length agreement, goal-potential improvement, goal entry, collision and clearance shaping, route adherence, forward motion, orientation consistency, the action-side flow-action bonus, and a navigation-side penalty for an executed premature STOP:
\begin{align}
    \widetilde{R}_c^{\mathrm{nav}}
    =\;&
    w_P P_c
    +
    w_L S_c^{\mathrm{len}}
    +
    w_G G_c
    +
    w_{\mathrm{entry}}I_c^{\mathrm{entry}}
    +
    R_{c,\mathrm{flow}}^a
    +
    R_{c,\mathrm{premature}}^{\mathrm{nav}}
    \nonumber\\
    &-
    w_{\mathrm{hard}}I_c^{\mathrm{hard}}
    -
    w_{\mathrm{repeat}}N_c^{\mathrm{repeat}}
    -
    w_{\mathrm{clear}}Q_c^{\mathrm{clear}}
    +
    w_{\mathrm{rec}}R_c^{\mathrm{rec}}
    \nonumber\\
    &-
    w_D D_c
    -
    w_B B_c
    -
    Y_c^{\mathrm{yaw}},
    \label{eq:app_navigation_reward_unclipped}
\end{align}
followed by
\begin{equation}
    R_c^{\mathrm{nav}}
    =
    \operatorname{clip}
    \left(
        \widetilde{R}_c^{\mathrm{nav}},
        -1,
        1
    \right).
    \label{eq:app_navigation_reward}
\end{equation}

The signed geodesic-progress term is
\begin{equation}
    P_c
    =
    \operatorname{clip}
    \left(
        \frac{
            d_c^s-d_c^e
        }{
            \max
            \left(
                L_{c,\mathrm{gt}}^{\mathrm{nav}},
                \delta_L
            \right)
        },
        -1,
        1
    \right),
    \label{eq:app_geodesic_progress}
\end{equation}
where $d_c^s$ and $d_c^e$ are the geodesic distances from the endpoints of the navigation-scored portion of chunk $c$ to the goal. This portion is truncated at the first goal-entry transition or an earlier execution termination. Accordingly, $L_{c,\mathrm{gt}}^{\mathrm{nav}}$ is the ground-truth path length over exactly these navigation-active transitions. The path-length agreement is
\begin{equation}
    S_c^{\mathrm{len}}
    =
    \exp
    \left[
        -
        \left(
            \frac{
                |L_c^{\mathrm{pred}}-L_c^{\mathrm{gt}}|
            }{
                \gamma_L L_c^{\mathrm{gt}}+\beta_L
            }
        \right)^2
    \right],
    \label{eq:app_path_length_agreement}
\end{equation}
where $L_c^{\mathrm{pred}}$ and $L_c^{\mathrm{gt}}$ are, respectively, the predicted and ground-truth path lengths over the full fixed horizon of $N_a$ transitions. In particular, $L_c^{\mathrm{pred}}$ includes all actions originally predicted for chunk $c$, including actions subsequently removed by STOP or collision resolution. Thus, $L_c^{\mathrm{gt}}$ differs from $L_{c,\mathrm{gt}}^{\mathrm{nav}}$ whenever navigation credit terminates before the end of the chunk; the two coincide otherwise. The goal-potential improvement is
\begin{equation}
    G_c
    =
    \Phi(d_c^e)-\Phi(d_c^s),
    \qquad
    \Phi(d)
    =
    \exp(-d/\tau_g).
    \label{eq:app_goal_potential}
\end{equation}

Let $j_{\mathrm{enter}}>0$ denote the first predicted position inside the goal radius. We set $I_c^{\mathrm{entry}}=1$ only for $c=1+\lfloor(j_{\mathrm{enter}}-1)/N_a\rfloor$; a trajectory initialized inside the goal region does not receive an entry bonus. Navigation credit in the entry chunk terminates at $j_{\mathrm{enter}}$, and subsequent executed transitions are assigned to the stopping stream.

\paragraph{Flow-action and premature-STOP terms.}
The gated flow-action score from Appendix~\ref{app:visual_reward} also provides action-side credit to the first chunk:
\begin{equation}
    R_{c,\mathrm{flow}}^a
    =
    \mathbf{1}[c=1]\,
    \lambda_{\mathrm{flow}}^a
    D_{\mathrm{deg}}
    S_{\mathrm{flow-act}},
    \qquad
    \lambda_{\mathrm{flow}}^a=0.05.
    \label{eq:app_action_flow_bonus}
\end{equation}
This term is set to zero when action decoding is invalid or when the first chunk is inactive. The flow-inferred yaw remains diagnostic only and does not enter $S_{\mathrm{flow-act}}$. An active chunk containing an actually executed incorrect STOP additionally receives
\begin{equation}
    R_{c,\mathrm{premature}}^{\mathrm{nav}}
    =
    -\left(
        \alpha_{\mathrm{nav}}
        +
        \beta_{\mathrm{nav}}
        q(d_{c,\mathrm{geo}}^{\mathrm{stop}})
    \right),
    \qquad
    \alpha_{\mathrm{nav}}=0.25,
    \quad
    \beta_{\mathrm{nav}}=0.75,
    \label{eq:app_premature_stop_navigation_penalty}
\end{equation}
and $R_{c,\mathrm{premature}}^{\mathrm{nav}}=0$ otherwise. Here, $q(\cdot)$ is the geodesic-distance scale defined in Eq.~\eqref{eq:app_premature_stop_scale}. This penalty is part of the navigation stream and is applied before the navigation reward is clipped.

\paragraph{Collision, clearance, and recovery shaping.}
The collision terms distinguish the raw generated plan from the deployment-simulated execution used for the first chunk. For $c=1$, let $n_1^{\mathrm{coll}}$ be the number of hard contacts encountered when executing the first chunk against the four-pixel-dilated occupancy map with last-free-on-segment clipping and continued execution after contact. The hard-contact indicator is
\begin{equation}
    I_c^{\mathrm{hard}}
    =
    \begin{cases}
        \mathbf{1}\!\left[n_1^{\mathrm{coll}}>0\right], & c=1,\\
        \mathbf{1}\!\left[\text{raw chunk $c$ collides under the 2-pixel margin}\right], & c>1.
    \end{cases}
    \label{eq:app_collision_hard}
\end{equation}
Repeated-contact credit is deployment-specific:
\begin{equation}
    N_c^{\mathrm{repeat}}
    =
    \begin{cases}
        \min\!\left(\max(n_1^{\mathrm{coll}}-1,0),2\right), & c=1,\\
        0, & c>1.
    \end{cases}
    \label{eq:app_collision_repeat}
\end{equation}
The two-pixel raw-plan collision check is also retained as a diagnostic for the first chunk.

Soft-clearance shaping is evaluated on the raw generated XY polyline of every active chunk, independently of the contact-clipped deployment execution. Let $\mathcal{P}_c$ denote the raster cells traversed by the raw polyline after excluding its fixed initial cell, and let $\kappa_{c,r}$ be the Euclidean obstacle clearance, in occupancy-map pixels, at cell $r\in\mathcal{P}_c$. With hard and soft margins $m_{\mathrm{hard}}=2$ px and $m_{\mathrm{soft}}=6$ px, the quadratic clearance risk is
\begin{equation}
    Q_c^{\mathrm{clear}}
    =
    \max_{r\in\mathcal{P}_c}
    \left[
        \operatorname{clip}
        \left(
            \frac{
                m_{\mathrm{soft}}-\kappa_{c,r}
            }{
                m_{\mathrm{soft}}-m_{\mathrm{hard}}
            },
            0,
            1
        \right)
    \right]^2.
    \label{eq:app_clearance_risk}
\end{equation}
Following the implementation, $Q_c^{\mathrm{clear}}=1$ if the raw path leaves the occupancy-map extent, and $Q_c^{\mathrm{clear}}=0$ if no raster cell beyond the fixed start is traversed.

Recovery credit is defined only for the deployment-simulated first chunk. After the last contact, let $e_1$ be the Euclidean distance from the contact-clipped position to the final executed position, and let $\kappa_1^{\mathrm{tail}}$ be the minimum raw-map clearance over the final two collision-free executed endpoints. A recovery is eligible only if the first chunk has at least one contact, is neither start-invalid nor out of bounds, has at least two collision-free actions after the last contact, satisfies $e_1\ge0.20\,\mathrm{m}$ and $\kappa_1^{\mathrm{tail}}\ge6$ px, and does not emit a post-collision premature STOP. For an eligible recovery,
\begin{equation}
    R_1^{\mathrm{rec}}
    =
    \min
    \left[
        \operatorname{clip}
        \left(
            \frac{e_1}{0.40\,\mathrm{m}},
            0,
            1
        \right),
        \operatorname{clip}
        \left(
            \frac{
                \kappa_1^{\mathrm{tail}}-4
            }{
                6-4
            },
            0,
            1
        \right)
    \right],
    \qquad
    R_c^{\mathrm{rec}}=0\;\text{for }c>1,
    \label{eq:app_collision_recovery_score}
\end{equation}
and $R_1^{\mathrm{rec}}=0$ otherwise. Thus, the first hard contact receives weight $w_{\mathrm{hard}}=0.60$, at most two additional contacts are penalized with weight $w_{\mathrm{repeat}}=0.10$ each, raw-path soft-clearance risk is weighted by $w_{\mathrm{clear}}=0.25$, and an eligible first-chunk recovery receives the bounded bonus $w_{\mathrm{rec}}R_1^{\mathrm{rec}}$ with $w_{\mathrm{rec}}=0.10$. These terms remain part of the navigation reward because the separate collision-credit stream is disabled.

\paragraph{Route and orientation terms.}
Let $\mathcal{I}_c$ denote the position indices of the navigation-active predicted trajectory supplied to the route diagnostic for chunk $c$, from the chunk start through its navigation endpoint inclusive. The endpoint is truncated at goal entry or an earlier execution termination; for the first chunk, the positions come from the four-pixel deployment-simulated trajectory, while later chunks use their corresponding predicted trajectories. If no navigation transition is active, the implementation sets $D_c=0$. Otherwise $n_c=|\mathcal{I}_c|\ge2$. For $i\in\mathcal{I}_c$, let $d_i^{\mathrm{route}}$ denote the distance from the predicted position to the nearest segment of the complete ground-truth route. The normalized route-deviation term is
\begin{equation}
    D_c=\operatorname{clip}\!\left(
    \frac{\sqrt{\frac{1}{n_c}\sum_{i\in\mathcal{I}_c}[(d_i^{\mathrm{route}}-0.75\,\mathrm{m})_+]^2}}
    {1.50\,\mathrm{m}},0,1\right).
    \label{eq:app_navigation_auxiliary_terms}
\end{equation}
For predicted steps longer than 0.03 m, $B_c$ is the mean of $\max(-\cos\angle(\Delta\mathbf{p}_i,\mathbf{t}_i^{\mathrm{GT}}),0)$, where $\mathbf{t}_i^{\mathrm{GT}}$ is the tangent of the ground-truth route segment nearest the predicted step midpoint. We set $B_c=0$ if no step satisfies the motion threshold. For the deployment-simulated first chunk, transitions marked by the collision-recovery grace mask are excluded from $B_c$ so that bounded recovery motion is not penalized as backward progress.

For orientation consistency, let $\theta_i^a$ be the cumulative predicted action yaw immediately before action $i$, $\theta_i^p$ the predicted-path heading relative to the first predicted XY step, and $\theta_i^{\mathrm{GT}}$ the corresponding ground-truth path heading. Both path-heading sequences are formed before applying the navigation aggregation masks. If the first XY step is numerically stationary ($\|\Delta\mathbf{p}_0\|_2<10^{-6}$), the canonical positive-X direction is used as the reference; subsequent steps with norm below $10^{-6}$ retain the most recent valid heading. The path-heading increment aligned with action $i$ is the forward difference
\begin{equation}
    \Delta\theta_i^p
    =
    \operatorname{wrap}_{[-\pi,\pi)}
    \left(
        \theta_{i+1}^p-\theta_i^p
    \right),
    \label{eq:app_path_heading_increment}
\end{equation}
which is computed in radians and has the same action index as $\Delta\psi_i$.

The final Stage-III run uses the linear-guard yaw score
\begin{equation}
    \ell(e;a,b)
    =
    \operatorname{clip}
    \left(
        \frac{|\operatorname{wrap}_{[-\pi,\pi)}(e)|_{\mathrm{deg}}-a}{b-a},
        0,
        1
    \right),
    \label{eq:app_yaw_linear_guard}
\end{equation}
where $|\cdot|_{\mathrm{deg}}$ converts the wrapped angular magnitude to degrees. Let $\mathcal{J}_c$ be the navigation-active action indices in chunk $c$ whose predicted XY translation satisfies $\|\Delta\mathbf{p}_i\|_2>0.03$ m. Path and rate consistency are evaluated only on $\mathcal{J}_c$. Their per-action errors are
\begin{equation}
    e_i^{\mathrm{path}}
    =
    \operatorname{wrap}_{[-\pi,\pi)}(\theta_i^a-\theta_i^p),
    \qquad
    e_i^{\mathrm{rate}}
    =
    \operatorname{wrap}_{[-\pi,\pi)}(\Delta\psi_i-\Delta\theta_i^p).
    \label{eq:app_yaw_path_rate_errors}
\end{equation}
For the collision-recovery grace transitions of the deployment-simulated first chunk, path consistency remains active but uses the folded path-error magnitude
\begin{equation}
    \bar e_i^{\mathrm{path}}
    =
    \min\!\left(
        |\operatorname{wrap}(e_i^{\mathrm{path}})|,
        \left|\pi-|\operatorname{wrap}(e_i^{\mathrm{path}})|\right|
    \right),
    \label{eq:app_yaw_recovery_fold}
\end{equation}
while the rate error is unchanged. Outside recovery grace, $\bar e_i^{\mathrm{path}}=|\operatorname{wrap}(e_i^{\mathrm{path}})|$. Hence the path and rate scores are
\begin{equation}
    z_i^{\mathrm{path}}
    =
    \ell(\bar e_i^{\mathrm{path}};10^\circ,45^\circ),
    \qquad
    z_i^{\mathrm{rate}}
    =
    \ell(e_i^{\mathrm{rate}};8^\circ,35^\circ).
    \label{eq:app_yaw_path_rate_scores}
\end{equation}

Gross-GT orientation credit additionally supervises pure-yaw controls. Let $\mathcal{K}_c$ be the navigation-active action indices satisfying either $\|\Delta\mathbf{p}_i\|_2>0.03$ m or $|\Delta\psi_i|>0.01$ rad, and let $\mathcal{G}_c$ denote the collision-recovery grace indices ($\mathcal{G}_c=\varnothing$ for chunks without deployment recovery). For a pure-yaw action, i.e., $\|\Delta\mathbf{p}_i\|_2\le0.03$ m and $|\Delta\psi_i|>0.01$ rad, the implementation evaluates the GT-heading error after applying the yaw increment; otherwise it evaluates the error before the action:
\begin{equation}
    e_i^{\mathrm{gross}}
    =
    \begin{cases}
        \operatorname{wrap}_{[-\pi,\pi)}(\theta_{i+1}^a-\theta_i^{\mathrm{GT}}),
        & \|\Delta\mathbf{p}_i\|_2\le0.03\,\mathrm{m},\ |\Delta\psi_i|>0.01\,\mathrm{rad},\\
        \operatorname{wrap}_{[-\pi,\pi)}(\theta_i^a-\theta_i^{\mathrm{GT}}),
        & \text{otherwise}.
    \end{cases}
    \label{eq:app_yaw_gross_error}
\end{equation}
The per-action gross score is $z_i^{\mathrm{gross}}=\ell(e_i^{\mathrm{gross}};12^\circ,60^\circ)$ and is aggregated only over $\mathcal{K}_c\setminus\mathcal{G}_c$; recovery-grace controls are therefore excluded from gross-GT orientation credit. Path and rate remain active on $\mathcal{J}_c$ during recovery grace, with only the path error folded as above.

Chunk-level yaw scores use the spike-sensitive mean--max aggregation
\begin{equation}
    \mathcal{A}_{\rho}\!\left(\{z_i\}_{i\in\mathcal{S}}\right)
    =
    (1-\rho)
    \frac{1}{|\mathcal{S}|}
    \sum_{i\in\mathcal{S}}z_i
    +
    \rho
    \max_{i\in\mathcal{S}}z_i,
    \qquad
    \rho=0.50,
    \label{eq:app_yaw_spike_aggregation}
\end{equation}
with value zero when $\mathcal{S}$ is empty. We then set
\begin{equation}
    \begin{aligned}
    Y_c^{\mathrm{path}}
    &=
    \mathcal{A}_{0.50}\!\left(\{z_i^{\mathrm{path}}\}_{i\in\mathcal{J}_c}\right), \\
    Y_c^{\mathrm{rate}}
    &=
    \mathcal{A}_{0.50}\!\left(\{z_i^{\mathrm{rate}}\}_{i\in\mathcal{J}_c}\right), \\
    Y_c^{\mathrm{gross}}
    &=
    \mathcal{A}_{0.50}\!\left(\{z_i^{\mathrm{gross}}\}_{i\in\mathcal{K}_c\setminus\mathcal{G}_c}\right).
    \end{aligned}
    \label{eq:app_yaw_chunk_scores}
\end{equation}
Finally, the three weighted yaw terms are capped jointly:
\begin{equation}
    Y_c^{\mathrm{yaw}}
    =
    \min
    \left(
        \kappa_{\mathrm{yaw}},
        w_{\mathrm{path}}Y_c^{\mathrm{path}}
        +
        w_{\mathrm{rate}}Y_c^{\mathrm{rate}}
        +
        w_{\mathrm{gross}}Y_c^{\mathrm{gross}}
    \right),
    \qquad
    \kappa_{\mathrm{yaw}}=0.24,
    \label{eq:app_yaw_total_penalty}
\end{equation}
with $w_{\mathrm{path}}=0.16$, $w_{\mathrm{rate}}=0.10$, and $w_{\mathrm{gross}}=0.08$. Equation~\eqref{eq:app_navigation_reward_unclipped} subtracts this non-negative capped penalty once, matching the implementation's negative yaw penalty with a lower bound of $-0.24$.

\input{tables/Appendix-Navigation-Reward}
\FloatBarrier

\subsubsection{Stopping Reward}
\label{app:stopping_reward}

Stage-III applies the ordered STOP detector independently to every chunk of $N_a=8$ actions. If the chunk's cumulative XY path length is strictly below $0.15$ m, STOP is emitted at the chunk-start position. Otherwise, the detector selects the first action satisfying $\max(|\Delta x_j|,|\Delta y_j|,|\Delta\psi_j|)\le10^{-3}$ and emits STOP immediately before that action is applied. A STOP in an earlier hypothetical chunk does not mask the independent STOP supervision of later chunks.

After the predicted trajectory enters the goal region, the stop-well term contains only a motion-energy penalty. Let $j_c^{\mathrm{end}}$ denote the exclusive upper transition index actually scored for chunk $c$ after STOP, collision, and other execution truncations are applied. Define
\begin{equation}
    E_j
    =
    \frac{
        (\|\mathbf{p}_{j+1}-\mathbf{p}_j\|_2-\delta_{xy})_+
    }{
        r_g
    }
    +
    w_\psi
    \frac{
        (|\Delta\psi_j|-\delta_\psi)_+
    }{
        \tau_\psi
    },
    \label{eq:app_action_energy}
\end{equation}
with $r_g=1.5$ m, $\delta_{xy}=0$, $w_\psi=0.18$, $\delta_\psi=0.02$ rad, and $\tau_\psi=0.25$ rad. The continuous stop-well reward is
\begin{equation}
    R_{c,\mathrm{well}}
    =
    -w_E
    \sum_{\substack{
        j=(c-1)N_a,\ldots,cN_a-1\\
        j\ge j_{\mathrm{enter}},\;j<j_c^{\mathrm{end}}
    }}
    E_j,
    \qquad
    w_E=0.20.
    \label{eq:app_stop_well_reward}
\end{equation}
Thus, unlike the navigation term $G_c$ in Eq.~\eqref{eq:app_goal_potential}, the final stopping stream contains no separate goal-potential improvement term.

For chunk $c$, let $\mathbf{p}_c^{\mathrm{start}}=\mathbf{p}_{(c-1)N_a}$ denote its start position. If the low-translation branch emits STOP, the stop position is exactly $\mathbf{p}_c^{\mathrm{start}}$. More generally, let $j_c^{\mathrm{stop}}$ denote the position index at which the ordered detector emits an executable STOP event and let $\mathbf{p}_c^{\mathrm{stop}}$ denote the corresponding position; for a componentwise near-zero action, this is the position immediately before that action is applied. STOP success is determined by Euclidean distance to the goal center,
\begin{equation}
    \|\mathbf{p}_c^{\mathrm{stop}}-\mathbf{g}\|_2\le r_g,
    \label{eq:app_stop_success_condition}
\end{equation}
whereas premature-STOP severity uses the occupancy-geodesic distance $d_{c,\mathrm{geo}}^{\mathrm{stop}}$ from the same stop position to the goal. Define
\begin{equation}
    q(d)
    =
    1-
    \exp
    \left[
        -
        \frac{(d-r_g-\delta_{\mathrm{fail}})_+}{\tau_{\mathrm{fail}}}
    \right],
    \qquad
    \delta_{\mathrm{fail}}=0.25\,\mathrm{m},
    \quad
    \tau_{\mathrm{fail}}=2.0\,\mathrm{m},
    \label{eq:app_premature_stop_scale}
\end{equation}
with $q(d)=1$ for a disconnected geodesic query. The hard STOP reward is
\begin{equation}
    R_{c,\mathrm{hard}}
    =
    \begin{cases}
        r_{\mathrm{succ}},
        & \text{an executable STOP is emitted and }\|\mathbf{p}_c^{\mathrm{stop}}-\mathbf{g}\|_2\le r_g,\\
        -\left(\alpha_{\mathrm{fail}}+\beta_{\mathrm{fail}}q(d_{c,\mathrm{geo}}^{\mathrm{stop}})\right),
        & \text{an executable STOP is emitted outside the goal region},\\
        0,
        & \text{no executable STOP is emitted},
    \end{cases}
    \label{eq:app_stop_hard_reward}
\end{equation}
where $r_{\mathrm{succ}}=0.15$, $\alpha_{\mathrm{fail}}=0.75$, and $\beta_{\mathrm{fail}}=0.25$, giving an incorrect-STOP reward in $[-1.00,-0.75]$. The same geodesic scale supplies the navigation-side premature-STOP term in Eq.~\eqref{eq:app_premature_stop_navigation_penalty}; that term uses $\alpha_{\mathrm{nav}}=0.25$ and $\beta_{\mathrm{nav}}=0.75$ but is not part of the stopping stream.

Leaving the goal region after entry incurs an indicator penalty $I_c^{\mathrm{exit}}$ with coefficient $w_{\mathrm{exit}}=1.0$. The complete stopping reward is
\begin{equation}
    R_c^{\mathrm{stop}}
    =
    \operatorname{clip}
    \left(
        R_{c,\mathrm{well}}
        +
        R_{c,\mathrm{hard}}
        -
        w_{\mathrm{exit}}I_c^{\mathrm{exit}},
        -1,
        1
    \right).
    \label{eq:app_stopping_reward}
\end{equation}

\input{tables/Appendix-Stopping-Reward}

\FloatBarrier

\subsubsection{Chunk Aggregation and Stage-III Objective}
\label{app:stage_three_aggregation}

We aggregate the chunk-wise navigation and stopping losses defined in
Eq.~\eqref{eq:stage3_action_loss} as
\begin{equation}
    \mathcal{L}_{\mathrm{nav}}
    =
    \sum_{c=1}^{B}
    \alpha_c
    \mathcal{L}_{\mathrm{nav},c},
    \qquad
    \sum_{c=1}^{B}\alpha_c=1,
    \label{eq:app_navigation_loss_aggregation}
\end{equation}
and
\begin{equation}
    \mathcal{L}_{\mathrm{stop}}
    =
    \frac{1}{B}
    \sum_{c=1}^{B}
    \mathcal{L}_{\mathrm{stop},c}.
    \label{eq:app_stopping_loss_aggregation}
\end{equation}

The navigation weights emphasize early chunks because receding-horizon
inference executes $\mathcal{A}_{t,1}$ before replanning. For $B=4$, we use
\begin{equation}
    (\alpha_1,\alpha_2,\alpha_3,\alpha_4)
    =
    \frac{(8,4,2,1)}{15}.
    \label{eq:app_navigation_chunk_weights}
\end{equation}

Substituting Eqs.~\eqref{eq:app_navigation_loss_aggregation} and~\eqref{eq:app_stopping_loss_aggregation} into Eq.~\eqref{eq:stage3_total_loss} gives the implemented Stage-III objective, which contains no auxiliary $\mathcal{L}_{\mathrm{WA}}$ term.

\FloatBarrier
\Needspace{0.24\textheight}

\subsubsection{Stage-III Optimization Settings}
\label{app:stage_three_optimization}

\input{tables/Appendix-StageIII-Optimization}

\FloatBarrier

\section{Benchmark Evaluation Details}
\label{app:evaluation_protocol}

\subsection{GN-Bench Closed-Loop Evaluation}
\label{app:gnbench_evaluation}

We evaluate every WNM checkpoint on the GN-Bench Seen and Unseen splits using the same receding-horizon inference procedure. The Seen and Unseen splits contain 1,000 and 5,000 episodes, respectively. The evaluator executes at most the first 8 actions before replanning. It emits STOP when the cumulative translational path length of this block, $\sum_{j=0}^{7}\|\Delta\mathbf{p}_j\|_2$, is below 0.15 m; an individual action whose three components have absolute value at most $10^{-3}$ also terminates the episode. These deployment checks are shared by WNM-2D and WNM-3D. Each checkpoint is evaluated once per episode with inference seed 1140. Navigation Error, Oracle Success, Success Rate, Trajectory Length, and SPL follow the benchmark definitions. We report point estimates over all episodes and do not compute multi-seed standard deviations or bootstrap confidence intervals.

\paragraph{Additional qualitative rollouts.}
\label{app:gnbench_qualitative}
Figure~\ref{fig:bn-bench-show} shows representative closed-loop WNM-3D rollouts on GN-Bench.

\input{figures/gn_bench}

\FloatBarrier

\subsection{VLN-CE Evaluation}
\label{app:vlnce_evaluation}

We evaluate WNM-2D and WNM-3D on the R2R Val-Unseen split of VLN-CE~\citep{krantz2020beyond}. For benchmark-specific training, both variants use the R2R~\citep{anderson2018vision} and RxR~\citep{ku2020room} training splits and follow the complete three-stage curriculum described in Sec.~\ref{sec:method}. In Stage-I, the dataset-provided reference paths are used as expert trajectories for supervised world-action learning. In Stage-II, corrective trajectories at policy-visited states are generated by a reference-aware navmesh oracle. When the current position is within 1\,m of the reference route, the oracle replans from the current position directly to the goal using Habitat-Sim's navmesh shortest-path query; otherwise, it first replans to a progress-aligned checkpoint on the reference route and then resumes the stored Stage-I reference trajectory. The implementation uses \texttt{habitat\_sim.ShortestPath} and \texttt{pathfinder.find\_path}; the resulting paths provide trajectory supervision rather than discrete actions from \texttt{ShortestPathFollower}. Stage-III then applies Counterfactual DanceGRPO to policy-specific rollouts. At inference, both variants receive only monocular egocentric RGB observations and language instructions. Each predicted action block contains eight local motion increments. A map-free kinematic beam-search controller tracks the predicted XY polyline using discrete turn and forward primitives, of which only the first four are executed before replanning; STOP is handled separately as described below. Episode horizon and other environment-side execution settings follow the official VLN-CE protocol. Neither model receives depth, camera pose, a bird's-eye-view observation, or an explicit metric map at inference time. Navigation Error, Oracle Success, Success Rate, and SPL follow the official benchmark definitions. Unless otherwise stated, optimizer, schedule, and Stage-III hyperparameter settings follow Appendices~\ref{app:sft_dagger_details} and~\ref{app:dancegrpo_details}; dataset sizes and rollout counts are benchmark-specific.

\paragraph{Action adaptation.}
WNM predicts eight local motion increments $(\Delta x,\Delta y,\Delta\psi)$ per action block. We accumulate the predicted $(\Delta x,\Delta y)$ components directly in the current query frame, without additional smoothing, to form a target XY polyline and track it with a map-free kinematic beam-search controller. Each expansion consists of zero or more same-direction turns followed by a $0.25$\,m forward move, with turn candidates spanning up to $\pm180^\circ$; turns are discretized at $15^\circ$ on R2R and $30^\circ$ on RxR. We use a beam width of 32 and search up to four forward moves. For a candidate containing $H\le4$ forward moves, the tracking cost is
\[
J=\sum_{h=0}^{H-1}0.7^h\left[
\frac{1}{5}\sum_{r=1}^{5}d_{h,r}^{2}
+d_{h,5}^{2}
+2\times10^{-5}|n_h|
-10^{-3}\Delta s_h
\right],
\]
where $d_{h,r}$ measures the distance to the target polyline at 20\%, 40\%, 60\%, 80\%, and 100\% of the $h$-th forward move, $n_h$ is the number of preceding discrete turns, and $\Delta s_h$ is the corresponding arc-length progress. Polyline association is constrained to move monotonically forward, with the admissible arc length capped by the starting progress plus the simulated travel distance and $0.015$\,m. The predicted yaw increment is not used in this tracking cost. The inference-time controller uses no map, navmesh query, obstacle cost, or ground-truth goal distance.

\paragraph{Execution and stopping.}
Only the first four discrete primitives of the selected sequence are executed before replanning; these may consist entirely of in-place turns. A collision during execution clears the remaining primitive queue and triggers an immediate model query. STOP is handled independently of beam search. The agent stops immediately if the cumulative translational path length of the predicted block is below $0.15$\,m or if all three components of the first predicted action have magnitude at most $10^{-3}$. A later near-zero action truncates the target motion prefix; STOP is appended only when the selected plan reaches that prefix endpoint with at most $0.1$\,m remaining arc length and $0.2$\,m terminal distance. Reaching the beam-search horizon alone does not trigger STOP.

\FloatBarrier
\Needspace{0.32\textheight}

\section{Real-World Deployment Details}
\label{app:real_world}

\paragraph{Deployment setup.}
We deploy WNM-3D on a laboratory-developed mobile robot equipped with an Intel RealSense D456 camera, from which only the monocular RGB stream is provided to the policy. Model inference runs on a remote GPU cluster as a server, while an NVIDIA Jetson Orin onboard the robot runs a ROS~2 client that sends navigation requests to the server through an SSH-tunneled TCP connection and receives the predicted action chunks for execution. A ROS~2 motion-execution node converts each predicted local pose increment $(\Delta x,\Delta y,\Delta\psi)$ into short-horizon base commands, which are tracked by the robot's low-level controller before the next replanning step. No additional onboard sensor signals are provided to WNM-3D. Action decoding and policy stopping follow Appendix~\ref{app:action_execution}.

\paragraph{Real-world adaptation.}
Starting from the simulation-trained Stage-III checkpoint, WNM-3D is further fine-tuned for 8,000 optimization steps on 210 self-collected real-world navigation trajectories using the supervised world-action objective $\mathcal{L}_{\mathrm{WA}}$ from Stage-I. We update the same trainable components as in Stage-I, namely the 3D Scene-to-Token Adapter and the world-action backbone, while keeping VGGT-$\Omega$ frozen. The adaptation scenes are disjoint from the scenes used in the qualitative real-world demonstrations.

%% file: tables/Appendix-Architecture.tex
\begin{table}[htbp]
    \caption{\textbf{Architecture configuration.} Shared world-action settings and WNM-3D-specific geometry and Adapter settings.}
    \label{tab:app_architecture_configuration}
    \centering
    \small
    \setlength{\tabcolsep}{5pt}
    \begin{NiceTabular}{c >{\raggedright\arraybackslash}p{0.54\linewidth} c}
        \toprule
        \textbf{Symbol} & \textbf{Configuration} & \textbf{Value} \\
        \midrule
        -- & World-action initialization & Wan2.2-TI2V-5B \\
        $K$ & History length & 33 \\
        -- & World-action input resolution & $160\times320$ \\
        -- & Geometry-encoder input resolution & $512\times512$ \\
        $B$ & Number of visual/action blocks & 4 \\
        $N_v$ & Future RGB frames per visual block & 8 \\
        $N_a$ & Actions per block & 8 \\
        -- & Selected VGGT-$\Omega$ encoder blocks & 5, 12, 18, 24 \\
        $L$ & Number of selected encoder blocks & 4 \\
        $d_s$ & Common adapter width & 512 \\
        $K\!\times\!H_s\!\times\!W_s$ & Source-memory lattice & $33\times32\times32$ \\
        $T_c\!\times\!H_c\!\times\!W_c$ & Target lattice & $9\times5\times10$ \\
        $N_c$ & History-prefix length (WNM-2D / WNM-3D) & 450 / 450 \\
        $R$ & Deformable resampling layers & 2 \\
        -- & Resampling heads / points per head & 8 / 8 \\
        -- & Spatiotemporal refinement blocks & 2 \\
        -- & Trainable adapter parameters & 23.97M \\
        \bottomrule
    \end{NiceTabular}
\end{table}

%% file: tables/Appendix-SFT-DAgger.tex
\begin{table}[htbp]
    \caption{\textbf{GN-Bench Stage-I and Stage-II training configuration.} Data and optimization settings for offline expert SFT and closed-loop DAgger-SFT.}
    \label{tab:app_sft_dagger_configuration}
    \centering
    \small
    \setlength{\tabcolsep}{6pt}
    \begin{NiceTabular}{lcc}
        \toprule
        \textbf{Configuration} & \textbf{Stage-I} & \textbf{Stage-II} \\
        \midrule
        Data source & A* demonstrations & Policy-specific DAgger \\
        Tasks / trajectories & 16K & 16K tasks \\
        WNM-2D chunks & -- & 691K \\
        WNM-3D chunks & -- & 633K \\
        Epochs & 20 & 5 \\
        Optimizer & AdamW & AdamW \\
        Initial learning rate & $1\times10^{-5}$ & $1\times10^{-5}$ \\
        Warmup ratio & $5\%$ & $5\%$ \\
        Weight decay & $1\times10^{-5}$ & $1\times10^{-5}$ \\
        Global batch: WNM-2D / WNM-3D & 256 / 192 & 256 / 192 \\
        GPUs & 16 H100 & 16 H100 \\
        \bottomrule
    \end{NiceTabular}
\end{table}

%% file: tables/Appendix-Visual-Reward.tex
\begin{table}[htbp]
    \caption{\textbf{Visual-reward configuration.} Parameters used for future-view quality and visual-action consistency in Stage-III.}
    \label{tab:app_visual_reward_configuration}
    \centering
    \small
    \setlength{\tabcolsep}{5pt}
    \begin{NiceTabular}{c >{\raggedright\arraybackslash}p{0.55\linewidth} c}
        \toprule
        \textbf{Symbol} & \textbf{Description} & \textbf{Value} \\
        \midrule
        $w_{\mathrm{pyr\text{-}ssim}}$ & Pyramid-SSIM weight & 0.60 \\
        $w_{\mathrm{charb}}$ & Charbonnier reconstruction weight & 0.25 \\
        $w_{\mathrm{temp}}$ & Temporal-consistency weight & 0.15 \\
        $\lambda_{\mathrm{flow}}^v$ & Visual flow-action weight & 0.05 \\
        $\delta_a$ & Action-motion threshold & 0.05 m \\
        $\delta_f$ & Flow-motion threshold & 0.01 m \\
        $\delta_{\mathrm{stat}}$ & Stationary-flow scale & 0.10 px \\
        -- & Flow input resolution & $64\times128$ \\
        -- & Flow spatial pooling grid & $4\times6$ \\
        -- & Flow descriptor statistics & Grid medians + p50/p75/p90 \\
        -- & Flow calibration samples & 480 ground-truth clips \\
        \bottomrule
    \end{NiceTabular}
\end{table}

%% file: tables/Appendix-Navigation-Reward.tex
\begin{table}[t]
    \caption{\textbf{Navigation-reward coefficients.} Coefficients, yaw-guard settings, flow-action credit, premature-STOP feedback, and collision-shaping settings used for Stage-III navigation credit.}
    \label{tab:app_navigation_reward_configuration}
    \centering
    \small
    \setlength{\tabcolsep}{3.5pt}
    \begin{NiceTabular}{c >{\raggedright\arraybackslash}p{0.22\linewidth} c @{\hspace{10pt}} c >{\raggedright\arraybackslash}p{0.22\linewidth} c}
        \toprule
        \textbf{Symbol} & \textbf{Description} & \textbf{Value} & \textbf{Symbol} & \textbf{Description} & \textbf{Value} \\
        \midrule
        $w_P$ & Geodesic-progress weight & 0.70 & $w_{\mathrm{path}}$ & Path-orientation weight & 0.16 \\
        $w_L$ & Path-length agreement weight & 0.15 & $w_{\mathrm{rate}}$ & Yaw-rate weight & 0.10 \\
        $w_G$ & Goal-potential weight & 0.15 & $w_{\mathrm{gross}}$ & Gross-GT yaw weight & 0.08 \\
        $w_{\mathrm{entry}}$ & Goal-entry bonus & 0.15 & $\kappa_{\mathrm{yaw}}$ & Total yaw-penalty cap & 0.24 \\
        $w_D$ & Route-deviation penalty & 0.35 & $\rho_{\mathrm{yaw}}$ & Mean--max mix & 0.50 \\
        $w_B$ & Backward-motion penalty & 0.30 & -- & Yaw score mode & linear guard \\
        $\delta_L$ & Progress normalization floor & 0.50m & $(a_{\mathrm{path}},b_{\mathrm{path}})$ & Path guard & $10^\circ/45^\circ$ \\
        $\gamma_L$ & Path-length scale & 0.50 & $(a_{\mathrm{rate}},b_{\mathrm{rate}})$ & Rate guard & $8^\circ/35^\circ$ \\
        $\beta_L$ & Path-length offset & 0.10m & $(a_{\mathrm{gross}},b_{\mathrm{gross}})$ & Gross-GT guard & $12^\circ/60^\circ$ \\
        $\tau_g$ & Goal-potential scale & 0.75m & $\delta_{\mathrm{xy}}^{\mathrm{yaw}}$ & Translation-motion floor & 0.03 m \\
        $\lambda_{\mathrm{flow}}^a$ & Action flow-action weight & 0.05 & $\delta_{\mathrm{rot}}^{\mathrm{yaw}}$ & Rotation-motion floor & 0.01 rad \\
        $\alpha_{\mathrm{nav}}$ & Premature-STOP nav base & 0.25 & $\beta_{\mathrm{nav}}$ & Premature-STOP nav add. & 0.75 \\
        $w_{\mathrm{hard}}$ & Hard-contact penalty & 0.60 & -- & Repeat-contact cap & 2 \\
        $w_{\mathrm{repeat}}$ & Repeat-contact penalty & 0.10 & -- & Hard diagnostic margin & 2 px \\
        $w_{\mathrm{clear}}$ & Soft-clearance penalty & 0.25 & -- & Deployment execution margin & 4 px \\
        $w_{\mathrm{rec}}$ & Recovery-bonus weight & 0.10 & -- & Recovery clearance threshold & 6 px \\
        -- & Soft-clearance margin & 6 px & -- & Recovery tail-free steps & 2 \\
        -- & Recovery min./full escape & 0.20 / 0.40 m & -- & -- & -- \\
        \bottomrule
    \end{NiceTabular}
\end{table}

%% file: tables/Appendix-Stopping-Reward.tex
\begin{table}[htbp]
    \caption{\textbf{Stopping-reward configuration.} Parameters used for goal settling, STOP detection, and distance-scaled premature-STOP supervision in Stage-III.}
    \label{tab:app_stopping_reward_configuration}
    \centering
    \small
    \setlength{\tabcolsep}{3.5pt}
    \begin{NiceTabular}{c >{\raggedright\arraybackslash}p{0.22\linewidth} c @{\hspace{10pt}} c >{\raggedright\arraybackslash}p{0.22\linewidth} c}
        \toprule
        \textbf{Symbol} & \textbf{Description} & \textbf{Value} & \textbf{Symbol} & \textbf{Description} & \textbf{Value} \\
        \midrule
        $r_g$ & Goal radius & 1.5 m & $\delta_{\mathrm{chunk}}$ & Chunk STOP threshold & 0.15 m \\
        $\delta_{xy}$ & Translation dead zone & 0 m & $\epsilon_{\mathrm{stop}}$ & Componentwise STOP threshold & $10^{-3}$ \\
        $w_\psi$ & Yaw-energy weight & 0.18 & $\delta_{\mathrm{fail}}$ & Premature-STOP deadband & 0.25 m \\
        $\delta_\psi$ & Yaw dead-zone threshold & 0.02 rad & $\tau_{\mathrm{fail}}$ & Premature-STOP distance scale & 2.0 m \\
        $\tau_\psi$ & Yaw normalization scale & 0.25 rad & $\alpha_{\mathrm{fail}}$ & Failure base magnitude & 0.75 \\
        $w_E$ & Action-energy weight & 0.20 & $\beta_{\mathrm{fail}}$ & Failure distance add. & 0.25 \\
        $r_{\mathrm{succ}}$ & STOP success reward & 0.15 & -- & Incorrect-STOP range & $[-1.00,-0.75]$ \\
        $w_{\mathrm{exit}}$ & Goal-exit penalty & 1.0 & -- & Stop-reward clip & $[-1,1]$ \\
        \bottomrule
    \end{NiceTabular}
\end{table}

%% file: tables/Appendix-StageIII-Optimization.tex
\begin{table}[htbp]
    \caption{\textbf{Stage-III optimization hyperparameters.} Main settings for Counterfactual DanceGRPO.}
    \label{tab:app_dancegrpo_hyperparameters}
    \centering
    \small
    \setlength{\tabcolsep}{3.5pt}
    \begin{NiceTabular}{c >{\raggedright\arraybackslash}p{0.24\linewidth} c @{\hspace{10pt}} c >{\raggedright\arraybackslash}p{0.24\linewidth} c}
        \toprule
        \textbf{Symbol} & \textbf{Setting} & \textbf{Value} & \textbf{Symbol} & \textbf{Setting} & \textbf{Value} \\
        \midrule
        -- & Global condition batch size & 64 & $\eta_v$ & Visual SDE coefficient & 0.70 \\
        $HK_s$ & Counterfactual branches / condition & 16 & $\eta_a$ & Action SDE coefficient & 0.20 \\
        -- & Sampler transitions & 16 & $\epsilon$ & Advantage stabilizer & $1\times10^{-4}$ \\
        -- & Eligible sampler transitions & 8 & $\varepsilon_v$ & Visual clipping threshold & $5\times10^{-4}$ \\
        -- & Replayed branch transitions / condition & 16 & $\varepsilon_a$ & Action clipping threshold & 0.01 \\
        -- & CFG scale & 5 & -- & Optimizer & AdamW \\
        -- & Learning rate & $5\times10^{-6}$ & -- & Stage-III updates & 700 \\
        $\lambda_{\mathrm{act}}$ & Action-side weight & 0.25 & $\lambda_{\mathrm{stop}}$ & Stopping weight & 0.50 \\
        $\{\alpha_c\}_{c=1}^{B}$ & Navigation chunk weights & $(8,4,2,1)/15$ & & & \\
        \bottomrule
    \end{NiceTabular}
\end{table}

%% file: figures/gn_bench.tex
\begin{figure}[htbp]
    \centering
    \includegraphics[width=\linewidth]{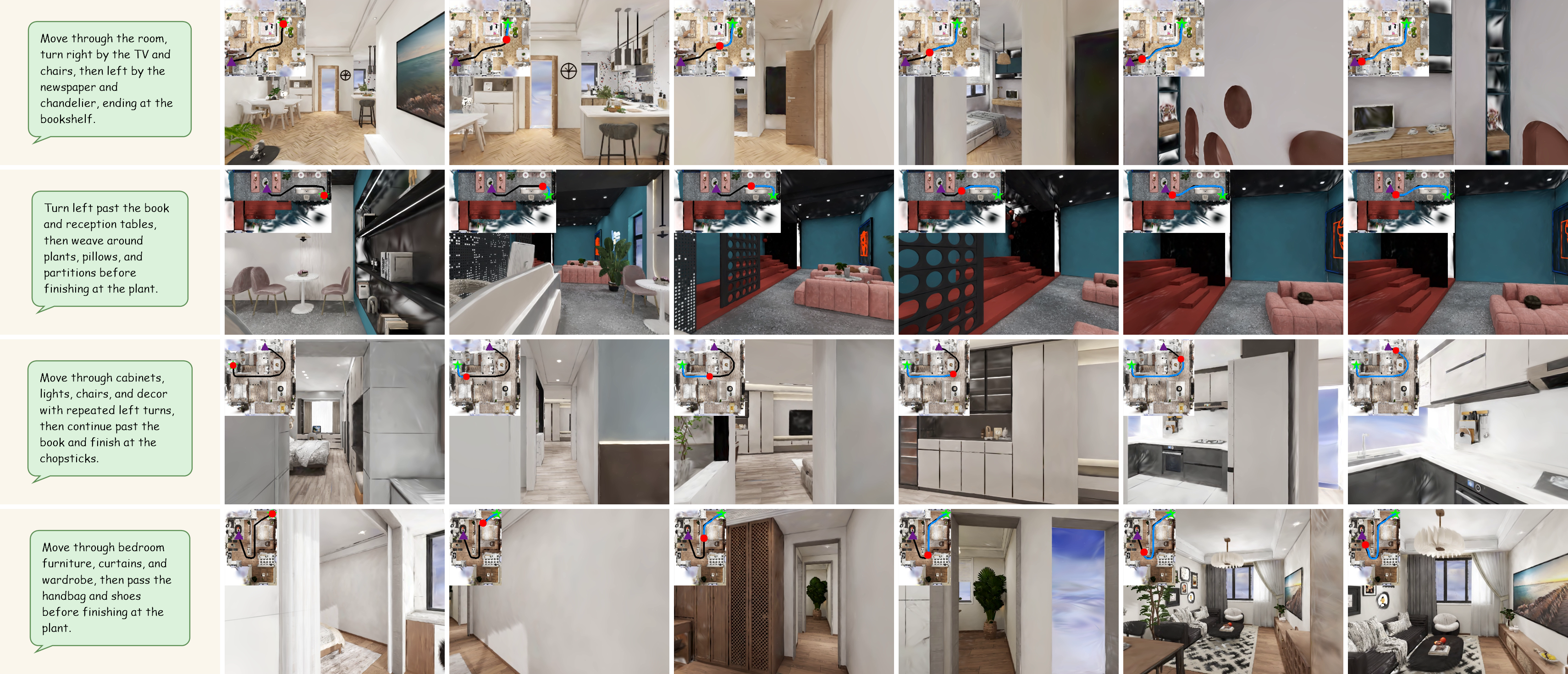}
    \caption{\textbf{Representative closed-loop WNM-3D rollouts on GN-Bench.} Each row shows a language instruction together with egocentric observations and the evolving navigation trajectory.}
    \label{fig:bn-bench-show}
\end{figure}